\documentclass[letterpaper]{article} 
\usepackage{aaai25}  
\usepackage{times}  
\usepackage{helvet}  
\usepackage{courier}  
\usepackage[hyphens]{url}  
\usepackage{graphicx} 
\usepackage{natbib}  
\usepackage{caption} 
\usepackage{algorithm}
\usepackage{algpseudocode}
\usepackage{multirow}
\usepackage{amsmath,amsthm,amsfonts,amssymb}
\allowdisplaybreaks
\usepackage{bm}
\usepackage{xcolor}

\theoremstyle{plain}
\newtheorem{thm}{Theorem}

\newtheorem{lem}{Lemma}

\usepackage{newfloat}
\usepackage{listings}
\DeclareCaptionStyle{ruled}{labelfont=normalfont,labelsep=colon,strut=off} 
\floatstyle{ruled}
\newfloat{listing}{tb}{lst}{}
\floatname{listing}{Listing}
\title{Federated t-SNE and UMAP for Distributed Data Visualization}
\author{
    Dong Qiao\equalcontrib,
    Xinxian Ma\equalcontrib,
    Jicong Fan\thanks{Corresponding author.}
}
\affiliations{
    School of Data Science, The Chinese University of Hong Kong, Shenzhen, China\\

    dongqiao@link.cuhk.edu.cn, 
xinxianma@link.cuhk.edu.cn, fanjicong@cuhk.edu.cn
}

\usepackage{bibentry}

\begin{document}

\maketitle

\begin{abstract}
High-dimensional data visualization is crucial in the big data era and these techniques such as t-SNE and UMAP have been widely used in science and engineering. Big data, however, is often distributed across multiple data centers and subject to security and privacy concerns, which leads to difficulties for the standard algorithms of t-SNE and UMAP. To tackle the challenge, this work proposes Fed-tSNE and Fed-UMAP, which provide high-dimensional data visualization under the framework of federated learning, without exchanging data across clients or sending data to the central server. The main idea of Fed-tSNE and Fed-UMAP is implicitly learning the distribution information of data in a manner of federated learning and then estimating the global distance matrix for t-SNE and UMAP. To further enhance the protection of data privacy, we propose Fed-tSNE+ and Fed-UMAP+. We also extend our idea to federated spectral clustering, yielding algorithms of clustering distributed data. In addition to these new algorithms, we offer theoretical guarantees of optimization convergence, distance and similarity estimation, and differential privacy. Experiments on multiple datasets demonstrate that, compared to the original algorithms, the accuracy drops of our federated algorithms are tiny.
\end{abstract}

%

\section{Introduction}

High-dimensional data are prevalent in science and engineering and their structures are often very complicated, which makes dimensionality reduction and data visualization appealing in knowledge discovery and decision-making \cite{jolliffe2016principal,hinton2006reducing,van2009dimensionality}. In the past decades, many algorithms have been proposed for dimensionality and visualization \cite{pearson1901liii,fisher1936use,sammon1969nonlinear,baker1977numerical,kohonen1982self,scholkopf1998nonlinear,roweis2000nonlinear,tenenbaum2000global,van2008visualizing,fan2018nonlinear,mcinnes2018umap}. 
Perhaps, the most popular algorithms in recent years are the t-distributed stochastic neighbor embedding (t-SNE) developed by \cite{van2008visualizing} and the Uniform Manifold Approximation and Projection (UMAP) proposed by \cite{mcinnes2018umap}. T-SNE and UMAP map the data points to a two- or three-dimensional space, exhibiting the intrinsic data distribution or pattern of the original high-dimensional data. Due to their superiority over other methods such as PCA \cite{jolliffe2016principal}, Isomap \cite{tenenbaum2000global}, and autoencoder \cite{hinton2006reducing}, they have been used for visualizing images, tabular data \cite{hao2021integrated}, text \cite{grootendorst2022bertopic}, and graphs \cite{NEURIPS2023_3ec6c6fc} in diverse fields and provide huge convenience for scientific research and engineering practice \cite{becht2019dimensionality}. Besides visualization, t-SNE and UMAP are also useful in clustering \cite{tsne_sdm} and outlier detection \cite{Fu_Zhang_Fan_2024}. There are also a few variants of t-SNE \cite{yang2009heavy,carreira2010elastic,xie2011m,van2014accelerating,gisbrecht2015parametric,pezzotti2016approximated,linderman2019fast,chatzimparmpas2020t,laptsne_2023} and UMAP \cite{sainburg2021parametric,nolet2021bringing}. For instance, \citet{van2014accelerating} used tree-based algorithms to accelerate the implementation of t-SNE. \citet{sainburg2021parametric} proposed a parametric UMAP that can visualize new data without re-training the model.

In many real cases such as mobile devices, IoT networks, medical records, and social media platforms, the high-dimensional data are distributed across multiple data centers and subject to security and privacy concerns \cite{dwork2014algorithmic,mcmahan2017communication,kairouz2021advances,qiao2024federated}, which leads to difficulties for the standard algorithms of t-SNE and UMAP. Specifically, in t-SNE and UMAP, we need to compute the pair-wise distance or similarity between all data points, meaning that different data centers or clients should share their data mutually or send their data to a common central server, which will leak data privacy and lose information security. To address this challenge, we propose federated t-SNE and federated UMAP in this work. Our main idea is implicitly learning the
distribution information of data in a manner of federated
learning and then estimating the global distance matrix for
t-SNE and UMAP. The contribution of this work is summarized as follows:
\begin{itemize}
    \item We propose Fed-tSNE and Fed-UMAP that are able to visualize distributed data of high-dimension.
    \item We further provide Fed-tSNE+ and Fed-UMAP+ to enhance privacy protection.
    \item We extend our idea to federated spectral clustering for distributed data with privacy protection.
    \item We provide theoretical guarantees such as reconstruction error bounds and differential privacy analysis.
\end{itemize}


\section{Related work}

\paragraph{t-SNE}
t-SNE \cite{van2008visualizing} aims to preserve the pair-wise similarities from high-dimension space $\mathcal{P}$ to low-dimension space $\mathcal{Q}$. The pair-wise similarities are measured as the probability that two data points are neighbors mutually. Specifically, given high-dimensional data points $\mathbf{x}_1, \mathbf{x}_2,\ldots,\mathbf{x}_N$ in $\mathbb{R}^{D}$, t-SNE computes the joint probability matrix $\mathbf{P}\in\mathbb{R}^{N\times N}$, in which $p_{ij} = 0$ if $i=j$, and 
$p_{i j}=\tfrac{p_{i \mid j}+p_{j \mid i}}{2 N}$,
if $i\neq j$, where 
\begin{equation}
\quad p_{j \mid i}=\tfrac{\exp \left(-\left\|\mathbf{x}_{i}-\mathbf{x}_{j}\right\|_{2}^{2} / (2 \tau_{i}^{2})\right)}{\sum_{\ell \in[N] \backslash\{i\}} \exp \left(-\left\|\mathbf{x}_{i}-\mathbf{x}_{\ell}\right\|_{2}^{2} / (2 \tau_{i}^{2})\right)}.\label{eq_pji}
\end{equation}
In \eqref{eq_pji}, $\tau_i$ is the bandwidth of the Gaussian kernel. Suppose $\mathbf{y}_1,\mathbf{y}_2,\ldots,\mathbf{y}_N$ are the low-dimensional embeddings in $\mathbb{R}^d$, where $d\ll D$,  t-SNE constructs a probability matrix $\mathbf{Q}$ by 
\begin{equation}
    q_{i j}=\tfrac{\left(1+\left\|\mathbf{y}_{i}-\mathbf{y}_{j}\right\|_{2}^{2}\right)^{-1}}{\sum_{\ell, s \in[N], \ell \neq s}\left(1+\left\|\mathbf{y}_{\ell}-\mathbf{y}_{s}\right\|_{2}^{2}\right)^{-1}}
\end{equation}
where $i\neq j$. Then t-SNE obtains $\mathbf{y}_1,\mathbf{y}_2,\ldots,\mathbf{y}_N$ by minimizing the Kullback-Leibler (KL) divergence
\begin{equation}
 \underset{\mathbf{y}_{1}, \ldots, \mathbf{y}_{N}}{\text{minimize}} ~\sum_{i \neq j} p_{i j} \log \frac{p_{i j}}{q_{i j}}
\label{f4}
\end{equation}

\paragraph{UMAP}
UMAP \cite{mcinnes2018umap} is a little similar to t-SNE. It starts by constructing a weighted k-NN graph in the high-dimensional space. The edge weights between points $\mathbf{x}_i$ and $\mathbf{x}_j$ are defined based on a fuzzy set membership, representing the probability that $\mathbf{x}_j$ is in the neighborhood of $\mathbf{x}_i$. Specifically, the membership strength is computed using
\begin{equation}
\mu_{i|j} = \exp\big(-{\|\mathbf{x}_i - \mathbf{x}_j\|^2}/{\sigma_i}\big),
\end{equation}
where $\sigma_i$ is a local scaling factor determined by the k-NNs of $\mathbf{x}_i$. The final membership strength is symmetrized as
\begin{equation}
\mu_{ij} = \mu_{i|j} + \mu_{j|i} - \mu_{i|j} \cdot \mu_{j|i}
\end{equation}
In the low-dimensional space, the probability of two points being neighbors is modeled using a smooth, differentiable approximation to a fuzzy set membership function. The edge weights between points $\mathbf{y}_i$ and $\mathbf{y}_j$ are given by
\begin{equation}
\mu'_{ij} = \tfrac{1}{1 + a \|\mathbf{y}_i - \mathbf{y}_j\|^{2b}}
\end{equation}
where $a$ and $b$ are hyperparameters typically set based on empirical data to control the spread of points in the low-dimensional space. UMAP minimizes the cross-entropy between the high-dimensional fuzzy simplicial set and the low-dimensional fuzzy simplicial set, i.e.,
\begin{equation}
\underset{\mathbf{y}_{1}, \ldots, \mathbf{y}_{N}}{\text{minimize}} ~ \sum_{i \neq j} \mu_{ij} \log \Big( \frac{\mu_{ij}}{\mu'_{ij}} \Big) + (1 - \mu_{ij}) \log \Big( \frac{1 - \mu_{ij}}{1 - \mu'_{ij}} \Big)
\end{equation}

\paragraph{Discussion}
Studies about federated dimensionality reduction or data visualization are scarce in the literature. \citet{grammenos2020federated} proposed a federated, asynchronous, and $(\epsilon,\delta)$-differentially private algorithm for PCA in the memory-limited setting. \citet{briguglio2023federated} developed a federated supervised PCA for supervised learning. \citet{novoa2023fast} proposed a privacy-preserving training algorithm for deep autoencoders. Different from PCA and autoencoders, in t-SNE and UMAP, we need to compute the pair-wise distance or similarity between data points, which leads to significantly greater difficulty in developing federated learning algorithms.
\citet{saha2022privacy} proposed a decentralized data stochastic neighbor embedding, dSNE. However, dSNE assumes that there is a shared subset of data among different clients, which may not hold in real applications.

\section{Federated Distribution Learning}
\subsection{Framework}
Suppose data $\mathbb R^{m\times n_x}\ni\bm X = \{\bm X_p\}_{p=1}^P$ are distributed at $P$ clients, where $\bm X_p\in\mathbb R^{m\times n_p}$ belongs to client $p$ and $\sum_{p = 1}^P n_p = n_x$. To implement t-SNE and UMAP, we need to compute a matrix $\bm{D}_{\bm{X},\bm{X}}\in\mathbb{R}^{n_x\times n_x}$ of distances between all data pairs in $\bm{X}$, which requires data sharing between the clients and central server, leading to data or privacy leaks. We propose to find an estimate of the distance or similarity matrix without data sharing. To do this, we let the central server construct a set of intermediate data points denoted by $\bm Y=[\bm y_1,\dots,\bm y_{n_y}]\in\mathbb R^{m\times n_y}$ and then compute distance matrices $\bm{D}_{\bm{Y},\bm{Y}}$ and $\{\bm{D}_{\bm{X}_p,\bm{Y}}\}_{p=1}^P$. These distance matrices can be used to construct an estimate $\widehat{\bm{D}}_{\bm{X},\bm{X}}$ of $\bm{D}_{\bm{X},\bm{X}}$ by applying the Nytr\"om method \cite{williams2000using} (to be detailed later). However, the choice of $\bm Y$ affects the accuracy of $\widehat{\bm{D}}_{\bm{X},\bm{X}}$, further influencing the performance of t-SNE and UMAP.

Since Nytr\"om method \cite{williams2000using} aims to estimate an entire matrix using its small sub-matrices, the sub-matrices should preserve the key information of the entire matrix, which means a good $\bm Y=[\bm y_1,\dots,\bm y_{n_y}]\in\mathbb R^{m\times n_y}$ should capture the distribution information of $\bm{X}$. Therefore, we propose to learn such a $\bm Y$ adaptively from the $P$ clients via solving the following federated distribution learning (FedDL) framework:
\begin{equation}\label{mdl:feddl}
\begin{aligned}
\mathop{\textup{minimize}}_{\bm Y}~F(\bm Y)\triangleq\sum_{p = 1}^P\omega_p f_p(\bm Y)
\end{aligned}
\end{equation}
where $f_p$ is the local objective function for each client, and $\omega_1,\dots,\omega_P$ are nonnegative weights for the clients. Without loss of generality, we set $\omega_1=\cdots=\omega_P=1/P$ for convenience in the remaining context. In this work, we set $f_p$ to be the Maximum Mean Discrepancy (MMD) \cite{gretton2012kernel} metric:
\begin{equation}
\begin{aligned}
& f_p(\bm Y) = \textup{MMD}(\bm X_p,\bm Y)\\
= & \frac{1}{n_p(n_p - 1)}\sum_{i=1}^{n_p}\sum_{j\neq i}^{n_p}k\left((\bm X_p)_{:,i},(\bm X_p)_{:,j}\right)\\
& -\frac{2}{n_p n_y}\sum_{i=1}^{n_p}\sum_{j=1}^{n_y}k\left((\bm X_p)_{:,i},(\bm Y)_{:,j}\right)\\
& + \frac{1}{n_y(n_y - 1)}\sum_{i=1}^{n_y}\sum_{j\neq i}^{n_y}k\left((\bm Y)_{:,i},(\bm Y)_{:,j}\right)
\end{aligned}
\end{equation}
or in the following compact form
\begin{equation}
\begin{aligned}
& f_p(\bm Y) = \textup{MMD}(\bm X_p,\bm Y)\\
= & \tfrac{1}{n_p(n_p - 1)}\left[\bm 1_{n_p}^T\bm{K}_{\bm X_p,\bm X_p}\bm 1_{n_p} - n_p\right]
 -\tfrac{2}{n_p n_y}\bm 1_{n_p}^T\bm{K}_{\bm X_p,\bm Y}\bm 1_{n_y}\\
& + \tfrac{1}{n_y(n_y - 1)}\left[\bm 1_{n_y}^T\bm{K}_{\bm Y,\bm Y}\bm 1_{n_y} - n_y\right]
\end{aligned}\label{f_p_mmd}
\end{equation}
where $k(\cdot,\cdot)$ is a kernel function and $\bm{K}_{\cdot,\cdot}$ denotes the kernel matrix computed from two matrices. MMD is a distance metric between two distributions and \eqref{f_p_mmd} is actually an estimation of MMD with finite samples from two distributions. If we use the Gaussian kernel $k(\bm x_i,\bm y_j) = \exp(-\gamma \|\bm x_i-\bm y_j\|^2)$, MMD compares all-order statistics between two distributions. For any $\bm X\in\mathbb R^{m\times n_x}$ and $\bm Y\in\mathbb R^{m\times n_y}$, we calculate the Gaussian kernel matrix as $\bm{K}_{\bm X,\bm Y} = \exp(-\gamma \bm{D}^2)$, where $\bm{D}^2$ is the squared pairwise distance matrix between $\bm X$ and $\bm Y$, i.e., $\bm{D}^2 = \textup{Diag}(\bm X^T\bm X)\bm 1_{n_y}^T - 2\bm X^T\bm Y + \bm 1_{n_x}\textup{Diag}(\bm Y^T\bm Y)^T$.


Combining \eqref{mdl:feddl} and \eqref{f_p_mmd}, we have the following optimization problem of federated distribution learning
\begin{equation}\label{mdl:feddl_mmd}
\begin{aligned}
\mathop{\textup{minimize}}_{\bm Y}~\sum_{p = 1}^P\omega_p \times \textup{MMD}(\bm X_p,\bm Y)
\end{aligned}
\end{equation}
By solving this problem, the central server or $\bm{Y}$ equivalently can learn the distribution information of the data distributed on the $P$ clients. Based on such an $\bm{Y}$, we can estimate the distance or similarity matrix between all data points in $\bm{X}$, which will be detailed later.


\begin{algorithm}[!h]
\caption{Federated Distribution Learning}\label{alg:feddl}
\begin{algorithmic}[1]
\Require Distributed data $\{\bm{X}_1,\bm{X}_2,\ldots,\bm{X}_P\}$ at $P$ clients.
\State Server broadcast an initial $\bm{Y}^0$ to all clients.
\For{round $s = 1$ to $S$}
\State \textbf{Client side:}
\For{client $p = 1$ to $P$ in parallel}
\State Set $\bm Y_p^{s, 0} = \bm Y^{s - 1}$
\State Update local variable $\bm Y_p^s$:
\For{$t = 1$ to $Q$}
\State $\bm Y_p^{s, t} = \bm Y_p^{s, t - 1} - \eta_s \nabla f_p(\bm Y_p^{s, t - 1})$
\EndFor
\State Denote $\bm Y_p^s = \bm Y_p^{s, Q}$
\State Upload $\bm Y_p^s$ (\textit{resp}., $\nabla f_p(\bm Y_p^{s,t})$) to the server.
\EndFor
\State \textbf{Server side:} compute
    $\bm Y^{s} = \frac{1}{{P}}\sum_{p=1}^P\bm Y_p^{s}$.
\State $\left(\textit{resp}., \bm Y^{s} \leftarrow \bm Y^{s - 1}-\eta_s'\times\frac{1}{{P}}\sum_{p=1}^P\nabla f_p(\bm{Y}_p^{s})\right)$
\State Broadcast $\bm Y^{s}$ to all clients.
\EndFor
\Ensure $\bm Y$
\end{algorithmic}
\end{algorithm}

\subsection{Optimization}
For a client $p$, we consider the corresponding local optimization problem
\begin{equation}
\begin{aligned}
\mathop{\textup{minimize}}_{\bm Y}~f_p(\bm Y)
\end{aligned}
\end{equation}
where $f_p(\bm Y) = \textup{MMD}(\bm X_p, \bm Y)$. 
Due to the presence of kernel function, we have to use some numerical methods like gradient descent to update the decision variable $\bm Y$. 
The gradient of $f_p$ at $\bm Y$ is
\begin{equation}
\begin{aligned}
\nabla f_p(\bm{Y}) = & \frac{-4\gamma}{n_p n_y}\left[\bm X_p\bm{K}_{\bm X_p,\bm Y} - \bm Y\textup{Diag}(\bm 1_{n_p}^T\bm{K}_{\bm X_p,\bm Y})\right]\\
 + & \frac{4\gamma}{n_y(n_y-1)}\left[\bm Y\bm{K}_{\bm Y,\bm Y} - \bm Y\textup{Diag}(\bm 1_{n_y}^T\bm{K}_{\bm Y,\bm Y})\right]
\end{aligned}\label{eq_fpY}
\end{equation}
To make it more explicit, we outline the key steps of FedDL to demonstrate how the central server coordinates local models for learning global distribution in a federated way.
\begin{itemize}
\item Step 1: The central server initializes a global $\bm Y_g$ before the learning cycle begins and broadcasts it to all participating local models.
\item Step 2: The local clients copy the global $\bm Y_g$ as their uniform initial guess $\bm{Y}_p$ and compute the gradient $\nabla f_p(\bm{Y}_p)$.
\item Step 3: Each client $p$ sends its gradient $\nabla f_p(\bm{Y}_p)$ or the updated $\bm{Y}$, i.e.,
\begin{align}
\bm Y_p \leftarrow \bm Y_p - \eta\nabla f_p(\bm{Y}_p)
\end{align}
to the central server,
where $\eta$ is the step size and can be set as the reverse of the Lipschitz constant of gradient if possible.
\item Step 4: The central server updates the global $\bm{Y}$ by averaging all posted $\bm Y_p$, i.e.,
\begin{equation}
    \bm Y = \frac{1}{{P}}\sum_{p=1}^P\bm Y_p,
\end{equation}
or performing gradient descent with the average of all $\nabla f_p(\bm{Y}_p)$, i.e.,
\begin{equation}
    \bm Y \leftarrow \bm Y-\eta'\times\frac{1}{{P}}\sum_{p=1}^P\nabla f_p(\bm{Y}_p),
\end{equation}
where $\eta'$ is a step size.
\item Step 5: The central server broadcasts the newly aggregated communication variables so as to trigger the next local updates.
\end{itemize}

The optimization details are summarized in Algorithm \ref{alg:feddl}. In the algorithm, for each client $p$, the time complexity per iteration is $\mathcal{O}(mn_p^2+mn_pn_y)$ and the space complexity is $\mathcal{O}(mn_p+mn_y+n_pn_y)$. 

In Algorithm \ref{alg:feddl}, it is necessary to share some variables like the global distribution information $\bm Y$ or the gradient $\nabla f_p(\bm Y)$ for proceeding the process of training. This may result in data privacy leakage. Data or gradient perturbation by some special types of noise is a common way to enhance the security of federated algorithms. In Section \ref{sec: feddl with dp}, we present the theoretical guarantees of distance estimation and similarity estimation and analyze the properties of differential privacy in such two ways, respectively.


\subsection{Convergence analysis}
Since we adopt MMD as our local objective function, they are all bounded below. Here, we give the convergence guarantee of Algorithm \ref{alg:feddl}.
\begin{thm}
Assume the gradient of all local objective functions $\{f_p\}_{p=1}^P$ are $L_p$-Lipschitz continuous, $L = \sum_{p=1}^P\omega_p L_p$ with $\omega_p = \frac{n_p}{n_x}$, $\rho_L = \frac{\sum_{p=1}^P\omega_p L_p^2}{L^2}$, and $\Vert\nabla f_p - \nabla f_{p'}\Vert_F \le \zeta$ for all $p,p'$, the sequence $\{\bm Y^{s,t}\}$ generated by Algorithm \ref{alg:feddl} with step size $1/L$ satisfies
\begin{equation}
\begin{aligned}
& \frac{1}{SQ}\sum_{s=1}^S\sum_{t=1}^Q\Vert\boldsymbol Y^{s,t} - \boldsymbol Y^{s,t-1}\Vert_F^2 \le \frac{4}{SQL}[F(\boldsymbol Y^0) - F(\boldsymbol Y^S)]\\
& + \frac{12\rho_L\zeta^2 (Q+1)(2Q+1)}{L^2[1 - 3(Q-1)^2(\rho_L + \frac{\max_p L_p^2}{L^2})]}\\
\end{aligned}
\end{equation}
\end{thm}
The proof can be found in Appendix \ref{prf: convergence of feddl}. It can be seen that when $SQ$ goes large enough, our algorithm converges to a finite value that is small provided that $\zeta$ is small. Figure \ref{fig:convergence} in Section \ref{sec_visualization} will show the convergence of the optimization numerically.


\begin{figure*}[t]
    \centering
    \includegraphics[width=1\textwidth]{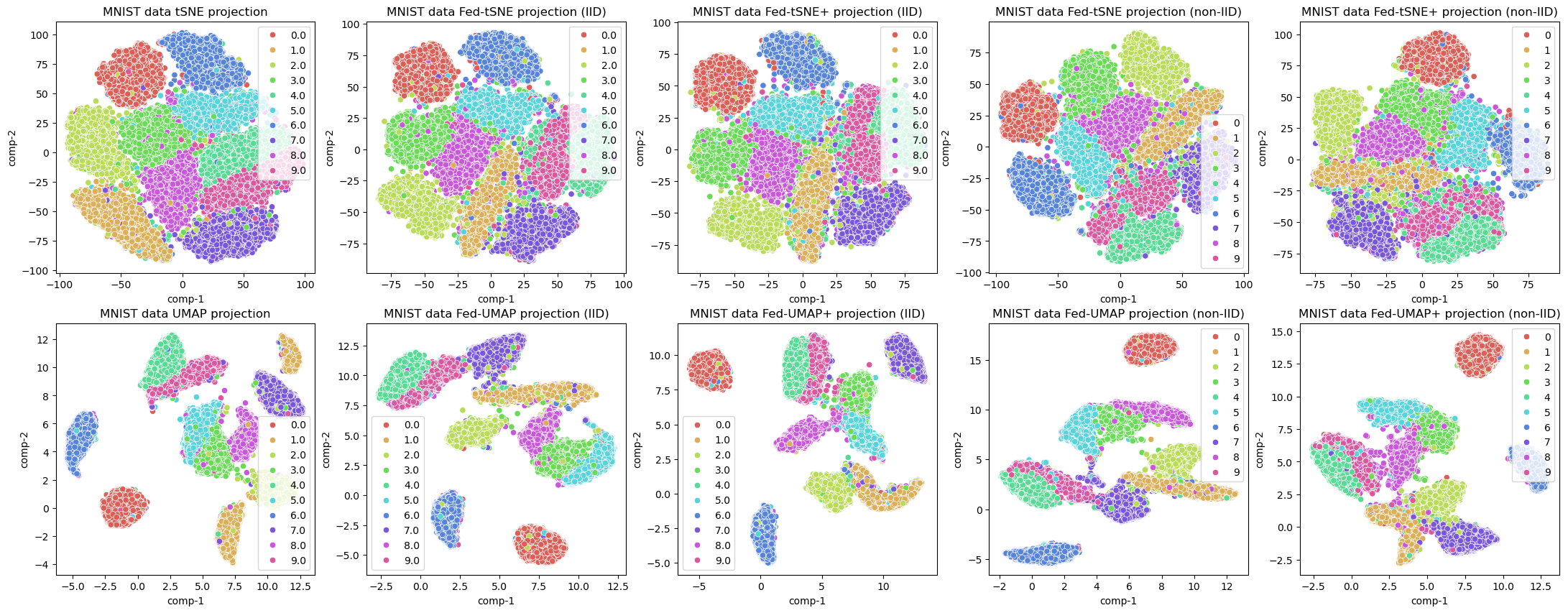}
    \caption{MNIST Data  Visualization. Row 1: t-SNE, Fed-tSNE, and Fed-tSNE+. Row 2: UMAP, Fed-UMAP, and Fed-UMAP+.}
    \label{fig:MNIST}
\end{figure*}

\section{Applications of FedDL}
\subsection{Federated tSNE and UMAP}
Nystrom approximation is a technique that can approximate a positive semi-definite (PSD) matrix merely through a subset of its rows and columns \cite{williams2000using}. Consider a PSD matrix $\mathcal{S}_+^n\ni\bm H\succeq 0$ that has a representation of block matrix
\begin{equation}
\begin{aligned}
\mathcal{S}_+^n\ni\bm H = \begin{bmatrix}
\bm W & \bm B^T\\
\bm B & \bm Z
\end{bmatrix}
\end{aligned}
\end{equation}
where $\bm W\in\mathcal{S}_+^c, \bm B\in\mathbb R^{(n-c)\times c}$, and $\bm Z\in\mathcal{S}_+^{n - c}$ for which $c\ll n$. Specifically, suppose $\bm{Z}$ is unknown, we can approximate it using $\bm W,\bm B$, and $\bm B^T$ as 
\begin{equation}
    \bm{Z}\approx \bm B\bm W_k^\dagger\bm B^T\triangleq \widehat{\bm{Z}}
\end{equation}
This means we can approximate the incomplete $\bm{H}$ by $\widehat{\bm{H}}=[\bm W, \bm B^T;\bm B, \widehat{\bm{Z}}]$.
By Nystr\"om method, we can approximate a distance or similarity matrix on large-scale dataset in a relatively low computational complexity. Some literature gives some useful upper bounds on Nystr\"om approximation in terms of Frobenius norm and spectral norm for different sampling techniques \cite{kumar2009sampling,JMLR:v6:drineas05a,zhang2008improved,kumar2009ensemble,li2010making}. Here, we present the upper bounds of Nystr\"om approximation in \cite{JMLR:v6:drineas05a} for our subsequent derivation.
\begin{thm}[Error bounds of Nystr\"om approximation]
Given $\bm X = [\bm x_1,\dots,\bm x_n]\in\mathbb R^{m\times n}$, let $\widehat{\bm H}$ be the rank-$k$ Nystrom approximation of $\bm H$ only through $c$ columns sampled uniformly at random without replacement from $\bm H$, and $\bm H_k$ be the best rank-$k$ approximation of $\bm H$. Then, the following inequalities hold for any sample of size $c$:
\begin{equation}
\begin{aligned}
&\Vert\bm H - \widehat{\bm H}\Vert_2 \le \Vert\bm H - \bm H_k\Vert_2 + \tfrac{2n\rho}{\sqrt{c}}\\
&\Vert\bm H - \widehat{\bm H}\Vert_F \le \Vert\bm H - \bm H_k\Vert_F + \rho\left(\tfrac{64k}{c}\right)^{1/4}\\
\end{aligned}
\end{equation}
where $\rho = \max_i \bm H_{ii}$.
\end{thm}

Without the retrieval of raw data from clients, we present federated tSNE (Fed-tSNE) and federated UMAP (FedUMAP) to visualize the high-dimensional data distributed across multiple regional centers. The main idea is to perform Algorithm \ref{alg:feddl} to learn a $\bm{Y}$ and then each client $p$ posts the distance matrix $\bm{D}_{\bm{X}_p,\bm{Y}}\in\mathbb{R}^{n_p\times n_y}$ between $\bm{X}_p$ and $\bm{Y}$ to the central server. Consequently, the central server assembles all $\bm{D}_{\bm{X}_p,\bm{Y}}$ to form 
\begin{equation}
    \bm{B}=[\bm{D}_{\bm{X}_1,\bm{Y}}^\top ~\bm{D}_{\bm{X}_2,\bm{Y}}^\top~ \cdots~ \bm{D}_{\bm{X}_P,\bm{Y}}^\top]^\top
    \label{eq_BD}
\end{equation}
and estimate $\bm{D}_{\bm{X},\bm{X}}$ as 
\begin{equation}
    \widehat{\bm{D}}_{\bm{X},\bm{X}}=\bm{B}\bm{W}_k^\dagger\bm{B}^\top
    \label{eq_Dxx}
\end{equation}
where $\bm{W}=\bm{D}_{\bm{Y},\bm{Y}}$, i.e., the distance matrix of $\bm{Y}$. Note that in the case that $\bm{W}$ is singular, we can add an identity matrix to it, i.e., $\bm{W}+\lambda \bm{I}$, where $\lambda>0$ is a small constant.
Finally, the central server implements either t-SNE or UMAP based on $\bm{D}_{\bm{X},\bm{X}}$. The steps are summarized into Algorithm \ref{alg_fedvisual}. 

\begin{algorithm}[!h]
\caption{Fed-tSNE and Fed-UMAP}\label{alg_fedvisual}
\begin{algorithmic}[1]
\Require Distributed data $\{\bm{X}_1,\bm{X}_2,\ldots,\bm{X}_P\}$ at $P$ clients.
\State Perform Algorithm \ref{alg:feddl} to compute $\bm{Y}$.
\State Each client $p$ computes the distance matrix $\bm{D}_{\bm{X}_p,\bm{Y}}$ and posts it to the central server.
\State The central server constructs $\bm{B}$ using \eqref{eq_BD} and computes $\widehat{\bm D}_{\bm{X},\bm{X}}$ using \eqref{eq_Dxx}.
\State The central server runs either t-SNE or UMAP on $\widehat{\bm D}_{\bm{X},\bm{X}}$ to obtain the low-dimensional embeddings $\bm{Z}$.
\Ensure $\bm{Z}$
\end{algorithmic}
\end{algorithm}

Note that sampling data points from clients like in classical Nystr\"om approximation is prohibitive in the federated settings. Thus, it motivates us to use FedDL to learn a useful set of fake points (\textit{i.e.}, landmarks) close enough to the data across the clients in terms of MMD.

\subsection{Federated Spectral Clustering}
Note that after running Algorithm \ref{alg:feddl}, if each client post the kernel matrix $\bm{K}_{\bm{X}_p,\bm{Y}}$ rather than the distance matrix $\bm{D}_{\bm{X}_p,\bm{Y}}$ to the central server, the central server can construct a kernel or similarity matrix $\widehat{\bm{K}}_{\bm{X},\bm{X}}$ that is useful for spectral clustering. Thus we obtain federated spectral clustering, of which the steps are summarized into Algorithm \ref{alg_fedsc}.
\begin{algorithm}[!h]
\caption{Fed-SpeClust}\label{alg_fedsc}
\begin{algorithmic}[1]
\Require Distributed data $\{\bm{X}_1,\bm{X}_2,\ldots,\bm{X}_P\}$ at $P$ clients.
\State Perform Algorithm \ref{alg:feddl} to compute $\bm{Y}$.
\State Each client $p$ computes the kernel matrix $\bm{K}_{\bm{X}_p,\bm{Y}}$ and posts it to the central server.
\State The central server constructs $\bm{C} =[\bm{K}_{\bm{X}_1,\bm{Y}}^\top ~\bm{K}_{\bm{X}_2,\bm{Y}}^\top~ \cdots~ \bm{K}_{\bm{X}_P,\bm{Y}}^\top]^\top$ and computes $\widehat{\bm{K}}_{\bm{X},\bm{X}} =\bm{C}\bm{W}^{-1}\bm{C}^\top$ with $\bm W =\bm{K}_{\bm Y, \bm Y}$.
\State The central server runs spectral clustering on $\widehat{\bm{K}}_{\bm{X},\bm{X}}$ to obtain the clusters $\mathcal{C}=\{\mathcal{C}_1,\mathcal{C}_2,\ldots,\mathcal{C}_c\}$.
\Ensure $\mathcal{C}$
\end{algorithmic}
\end{algorithm}

\section{FedDL with differential privacy}
\label{sec: feddl with dp}
\subsection{FedDL by data perturbation}
We inject noise into the raw data in each client and then run FedDL to learn the global distribution information. Note that data perturbation is a one-shot operation before performing Algorithm \ref{alg:feddl}. Specifically, the data $\bm X$ is perturbed by a noise matrix $\bm{E}\in\mathbb{R}^{m\times n_x}$ to form the noisy data matrix
$\label{mdl:feddl with perturbed data}
\Tilde{\bm X} = \bm X + \bm E,
$
where $e_{i,j}\sim \mathcal{N}(0,\sigma^2)$. Define $\Tilde{\bm X} = \{\Tilde{\bm X_p}\}_{p=1}^P$ and we then perform Algorithm \ref{alg:feddl} on $\Tilde{\bm X}$ to obtain $\bm Y$ which gives the Nystr\"om approximation
\begin{equation}
\begin{aligned}
\widehat{\bm H}_{\Tilde{\bm X}, \Tilde{\bm X}|\bm Y} \simeq \bm B\bm W_k^\dagger\bm B^T
\end{aligned}
\end{equation}
where $\bm B = \bm{K}_{\Tilde{\bm X}, \bm Y}$ (or $\bm{D}_{\Tilde{\bm X},\bm{Y}}$), $\bm W = \bm{K}_{\bm Y, \bm Y}$ (or $\bm{D}_{\bm Y, \bm Y}$).

Following the logistics of existing literature, we give the upper bounds on the approximation error of Nystr\"om approximation involved with FedDL, where we focus only on the kernel matrix because it is more complex than the distance matrix.
\begin{thm}[Error bound of Nystr\"om approximation with FedDL having data perturbation]\label{thm:reconstruction error 1}
Given $\bm X = \{\bm X_p\}_{p=1}^P$ with $\bm X_p\in\mathbb R^{m\times n_p}$ having $\sum_{p = 1}^P n_p = n_x$, $\bm Y=[\bm y_1,\dots,\bm y_{n_y}]\in\mathbb R^{m\times n_y}$, let $\Tilde{\bm X}_a^x = [\bm Y, \Tilde{\bm X}]$ be the augmented matrix, $\bm C = \bm K_{\Tilde{\bm X}_a^x,\bm Y}$, $\bm W = \bm K_{\bm Y,\bm Y}$ with $\bm W_k^\dagger$ being the Moore-Penrose inverse of the best rank-$k$ approximation of $\bm W$, $\xi_m = \sqrt{m+\sqrt{2mt}+2t}$, and $\text{Cond}(\cdot)$ denote condition number of matrix. Denoting $\widehat{\bm H}_{\Tilde{\bm X},\Tilde{\bm X}|\bm Y} = \bm C\bm W_k^\dagger\bm C^T$, it holds with probability at least $1 - n(n-1)e^{-t}$ that
\begin{equation*}
\begin{aligned}
&\Vert\widehat{\bm H}_{\Tilde{\bm X}, \Tilde{\bm X}|\bm Y} - \bm{K}_{\bm X,\bm X}\Vert_2\\
\le & \textup{Cond}(\bm{K}_{\Tilde{\bm X}_a^x,\Tilde{\bm X}_a^x})\left(\tfrac{\mid\textup{MMD}(\Tilde{\bm X},\bm Y)\mid}{n_x + n_y} + 1\right) + 2n_x\\
& + \sqrt{2}n_x\gamma\left[\sigma^2\xi_m^2 + \sqrt{2}{\left\Vert \bm{D}_{\bm X,\bm X}\right\Vert_\infty}\sigma\xi_m\right]
\end{aligned}
\end{equation*}
alternatively, it holds that
\begin{equation*}
\begin{aligned}
&\big\Vert\widehat{\bm H}_{\Tilde{\bm X}, \Tilde{\bm X}|\bm Y} - \bm{K}_{\bm X, \bm X}\big\Vert_F\\
\le & \sqrt{n_x + n_y - k}\textup{Cond}(\bm{K}_{\Tilde{\bm X}_a^x,\Tilde{\bm X}_a^x})\left(\tfrac{\mid\textup{MMD}(\Tilde{\bm X},\bm Y)\mid}{n_x + n_y} + 1\right)\\
+& 2k^{\frac{1}{4}}n_x\sqrt{1 + \tfrac{n_y}{n_x}}
 + \sqrt{2}n_x\gamma\left[\sigma^2\xi_m^2 + \sqrt{2}{\left\Vert \bm{D}_{\bm X,\bm X}\right\Vert_\infty}\sigma\xi_m\right]
\end{aligned}
\end{equation*}
\end{thm}

\begin{thm}[Differential privacy of FedDL with data perturbation]\label{thm:dp1}
Assume $\max_{p,j}\Vert(\bm X_p)_{:,j}\Vert_2 = \tau_X$, FedDL with perturbed data given by Section \ref{mdl:feddl with perturbed data} is $(\varepsilon,\delta)-$differentially private if $\delta\ge 2c\tau_X/\varepsilon$, where $c^2 > 2\ln(1.25/\delta)$.
\end{thm}

\subsection{FedDL by variable and gradient perturbation}
We can also perturb the optimization variable $\bm Y$ or the gradient $\nabla f_p(\bm Y_p)$ by Gaussian noise in the training progression to improve the security of Algorithm \ref{alg:feddl}. No matter which method we follow, the $\bm{Y}$ obtained by the central server is noisy, i.e., $\Tilde{\bm Y} = \bm Y + \bm E$, where $\bm E$ is drawn elementwise from $\mathcal{N}(0,\sigma^2)$. Then, we do Nystrom approximation by
\begin{equation*}
\begin{aligned}
\widehat{\bm{H}}_{\bm X, \bm X|\Tilde{\bm Y}} \simeq \bm B\bm W_k^\dagger\bm B^T
\end{aligned}
\end{equation*}
where $\bm B = \bm{K}_{\bm X, \Tilde{\bm Y}}$ (or $\bm{D}_{\bm X, \Tilde{\bm Y}}$), $\bm W = \bm{K}_{\Tilde{\bm Y}, \Tilde{\bm Y}}$ (or $\bm{D}_{\Tilde{\bm Y}, \Tilde{\bm Y}}$).

\begin{thm}[Error bound of Nystr\"om approximation with FedDL having gradient perturbation]\label{thm:reconstruction error 2}
With the same notations in Theorem \ref{thm:reconstruction error 1}, let $\Tilde{\bm X}_a^y = [\Tilde{\bm Y}, \bm X]$ be the augmented matrix. Then it holds that
\begin{equation*}
\begin{aligned}
&\big\Vert\widehat{\bm{H}}_{\bm X, \bm X|\Tilde{\bm Y}} - \bm{K}_{\bm X, \bm X}\big\Vert_2\\
\le &\textup{Cond}(\bm{K}_{\Tilde{\bm X}_a^y,\Tilde{\bm X}_a^y})\left(\tfrac{\mid\textup{MMD}(\bm X, \Tilde{\bm Y})\mid}{n_x + n_y} + 1\right) + 2n_x
\end{aligned}
\end{equation*}
alternatively, it holds that
\begin{equation*}
\begin{aligned}
&\big\Vert\widehat{\bm{H}}_{\bm X, \bm X|\Tilde{\bm Y}} - \bm{K}_{\bm X, \bm X}\big\Vert_F\\
\le & \sqrt{n_x + n_y - k}\textup{Cond}\left(\bm{K}_{\Tilde{\bm X}_a^y,\Tilde{\bm X}_a^y}\right)\left(\tfrac{\mid\textup{MMD}(\bm X,\Tilde{\bm Y})\mid}{n_x + n_y} + 1\right)\\
& + 2k^{1/4}n_x\sqrt{1 + \tfrac{n_y}{n_x}}
\end{aligned}
\end{equation*}
\end{thm}
Note that $\textup{MMD}(\bm X,\Tilde{\bm Y})\leq \textup{MMD}(\bm X,{\bm Y})+\textup{MMD}(\bm Y,\Tilde{\bm Y})$ is related to $\sigma$. A smaller $\sigma$ leads to a lower estimation error (higher estimation accuracy) but weaker privacy protection. We can obtain a precise trade-off between accuracy and privacy by combining Theorem \ref{thm:reconstruction error 2} with Theorem \ref{thm:dp2}.

\begin{table*}[!h]
\centering
\begin{tabular}{l||c|c|c|c|c}
\hline
        & & \multicolumn{2}{c|}{IID}  & \multicolumn{2}{c}{non-IID} \\ \hline \hline
Metric & tSNE & Fed-tSNE & Fed-tSNE+ &  Fed-tSNE & Fed-tSNE+ \\ \hline
CA 1-NN  & 0.9618$\pm$0.0015     & 0.9400$\pm$0.0017         & 0.9364$\pm$0.0020     & 0.9412$\pm$0.0021     & 0.9189$\pm$0.0030                \\
CA 10-NN  & 0.9656$\pm$0.0017     & 0.9477$\pm$0.0017         & 0.9443$\pm$0.0012     & 0.9483$\pm$0.0019     & 0.9307$\pm$0.0026              \\
CA 50-NN  & 0.9609$\pm$0.0015     & 0.9401$\pm$0.0022         & 0.9354$\pm$0.0022     & 0.9406$\pm$0.0020     & 0.9209$\pm$0.0035              \\
NPA 1-NN  & 0.4176$\pm$0.0016     &  0.2728$\pm$0.0022        & 0.2543$\pm$0.0016     & 0.2729$\pm$0.0022     & 0.1928$\pm$0.0019               \\
NPA 10-NN  & 0.3905$\pm$0.0005     & 0.3373$\pm$0.0007         & 0.3263$\pm$0.0005     & 0.3375$\pm$0.0013     & 0.2827$\pm$0.0010               \\
NPA 50-NN  & 0.3441$\pm$0.0007     & 0.3301$\pm$0.0007         & 0.3258$\pm$0.0007     & 0.3305$\pm$0.0006     & 
0.3030$\pm$0.0012\\
NMI    & 0.7747$\pm$0.0243     &  0.7534$\pm$0.0202        &  0.7471$\pm$0.0073    & 0.7399$\pm$0.0109     &  0.7025$\pm$0.0149             \\
SC     & 0.4226$\pm$0.0082     & 0.4407$\pm$0.0103         & 0.4478$\pm$0.0066     & 0.4321$\pm$0.0058     & 0.4441$\pm$0.0045              \\ \hline \hline
Metric & UMAP & Fed-UMAP & Fed-UMAP+ &  Fed-UMAP & Fed-UMAP+ \\ \hline
CA 1-NN  & 0.9322$\pm$0.0053     &  0.9066$\pm$0.0031        & 0.9007$\pm$0.0034     &  0.9064$\pm$0.0026    &   0.8730$\pm$0.0041              \\
CA 10-NN  &  0.9613$\pm$0.0048     &  0.9445$\pm$0.0018        & 0.9416$\pm$0.0023     & 0.9449$\pm$0.0022     & 0.9224$\pm$0.0036              \\
CA 50-NN  &  0.9602$\pm$0.0049    &  0.9432$\pm$0.0020        & 0.9400$\pm$0.0025     & 0.9441$\pm$0.0022     & 0.9219$\pm$0.0037              \\
NPA 1-NN  &  0.0308$\pm$0.0009    &  0.0293$\pm$0.0007        & 0.0277$\pm$0.0008     &  0.0298$\pm$0.0011    & 0.0218$\pm$0.0009               \\
NPA 10-NN  & 0.1227$\pm$0.0010     &  0.1133$\pm$0.0008        & 0.1088$\pm$0.0009     & 0.1131$\pm$0.0012     & 0.0914$\pm$0.0006               \\
NPA 50-NN  & 0.2226$\pm$0.0015     &  0.2099$\pm$0.0011        & 0.2053$\pm$0.0011     &  0.2095$\pm$0.0013    & 0.1860$\pm$0.0013               \\
NMI    & 0.8285$\pm$0.0150     &  0.7844$\pm$0.0208        & 0.7812$\pm$0.0153     &  0.7919$\pm$0.0217    & 0.7368$\pm$0.0194              \\
SC     & 0.6118$\pm$0.0207     &  0.5812$\pm$0.0261        & 0.5746$\pm$0.0229     & 0.5889$\pm$0.0248     & 0.5422$\pm$0.0173              \\ \hline \hline
\end{tabular}
\caption{Performance (mean$\pm$std) of dimensionality reduction on MNIST}
\label{table-mnist}
\end{table*}

\begin{table*}[!h]
\centering
\begin{tabular}{l||c|c|c|c|c}
\hline
        & & \multicolumn{2}{c|}{IID}  & \multicolumn{2}{c}{non-IID} \\ \hline \hline
Metric & tSNE & Fed-tSNE & Fed-tSNE+ &  Fed-tSNE & Fed-tSNE+ \\ \hline
CA 1-NN  & 0.8112$\pm$0.0049     & 0.7473$\pm$0.0029         & 0.7198$\pm$0.0041     & 0.7453$\pm$0.0044     & 0.6669$\pm$0.0044                \\
CA 10-NN  & 0.8260$\pm$0.0039     & 0.7892$\pm$0.0030         & 0.7706$\pm$0.0034     & 0.7898$\pm$0.0039     & 0.7280$\pm$0.0048              \\
CA 50-NN  & 0.8064$\pm$0.0041     & 0.7754$\pm$0.0033         & 0.7631$\pm$0.0037     & 0.7760$\pm$0.0045     & 0.7280$\pm$0.0043              \\
NPA 1-NN  & 0.3518$\pm$0.0018     &  0.1251$\pm$0.0021        & 0.0718$\pm$0.0013     & 0.1275$\pm$0.0017     & 0.0274$\pm$0.0006               \\
NPA 10-NN  & 0.3635$\pm$0.0007     & 0.2551$\pm$0.0010         & 0.1954$\pm$0.0011     & 0.2571$\pm$0.0011     & 0.1090$\pm$0.0010               \\
NPA 50-NN  & 0.3710$\pm$0.0003     & 0.3363$\pm$0.0006         & 0.3004$\pm$0.0006     & 0.3369$\pm$0.0008     & 0.2204$\pm$0.0017\\
NMI    & 0.5787$\pm$0.0212     &  0.5780$\pm$0.0154        &  0.5733$\pm$0.0149    & 0.5778$\pm$0.0044     &  0.5162$\pm$0.0129             \\
SC     & 0.4049$\pm$0.0101     & 0.4382$\pm$0.0070         & 0.4638$\pm$0.0147     & 0.4389$\pm$0.0085     & 0.4564$\pm$0.0111              \\ \hline \hline
Metric & UMAP & Fed-UMAP & Fed-UMAP+ &  Fed-UMAP & Fed-UMAP+ \\ \hline
CA 1-NN  & 0.7146$\pm$0.0029     &  0.6756$\pm$0.0036        & 0.6587$\pm$0.0055     &  0.6766$\pm$0.0043    &   0.6110$\pm$0.0037              \\
CA 10-NN  &  0.7734$\pm$0.0039     &  0.7413$\pm$0.0045        & 0.7287$\pm$0.0041     & 0.7437$\pm$0.0030     & 0.6875$\pm$0.0041              \\
CA 50-NN  &  0.7781$\pm$0.0039    &  0.7491$\pm$0.0052        & 0.7383$\pm$0.0039     & 0.7501$\pm$0.0040     & 0.7006$\pm$0.0033              \\
NPA 1-NN  &  0.0356$\pm$0.0012    &  0.0218$\pm$0.0011        & 0.0156$\pm$0.0009     &  0.0223$\pm$0.0011    & 0.0071$\pm$0.0004               \\
NPA 10-NN  & 0.1401$\pm$0.0013     &  0.1002$\pm$0.0015        & 0.0799$\pm$0.0012     & 0.1020$\pm$0.0010     & 0.0423$\pm$0.0007               \\
NPA 50-NN  & 0.2518$\pm$0.0018     &  0.2152$\pm$0.0028        & 0.1907$\pm$0.0018     &  0.2167$\pm$0.0022    & 0.1226$\pm$0.0015               \\
NMI    & 0.6187$\pm$0.0127     &  0.5915$\pm$0.0112        & 0.5755$\pm$0.0090     &  0.5877$\pm$0.0181    & 0.5191$\pm$0.0132              \\
SC     & 0.5304$\pm$0.0286     &  0.5448$\pm$0.0264        & 0.5476$\pm$0.0176     & 0.5338$\pm$0.0252     & 0.5322$\pm$0.0191              \\ \hline \hline
\end{tabular}
\caption{Performance (mean$\pm$std) of dimensionality reduction on Fashion-MNIST}
\label{table-fmnist}
\end{table*}

\begin{thm}[Differential privacy of FedDL with gradient perturbation]\label{thm:dp2}
Suppose $\max_{p,j}\Vert(\bm X_p)_{:,j}\Vert_2 = \tau_X$, $\max_{p,i,j}\Vert(\bm Y_{p})_{:,i} - (\bm X_p)_{:,j}\Vert = \Upsilon$, $\Vert\bm Y_p^s\Vert_{sp} \le \tau_Y \forall s$, let $\{\nabla f_p(\bm Y_p^s)\}_{p = 1}^P$ for $s\in[S]$ be the sequence that is perturbed by noise drawn from $\mathcal{N}(0,\sigma^2)$ with variance $8S\Delta^2\log(e + (\varepsilon/\delta))/\varepsilon^2$ where $\Delta = \frac{8\sqrt{n_y}\gamma\tau_X}{n_pn_y}\left\{1 + 2\gamma(\tau_X + \tau_Y)\left(\tau_X + \Upsilon\right)\right\}$. Then, the Gaussian Mechanism that injects noise to $\{\nabla f_p(\bm Y_p^s)\}_{s = 1}^S$ for $p\in[P]$ is $(\varepsilon,\delta)-$differentially private.
\end{thm}

Note that it is intuitively appropriate to choose a decreasing sequence of noise variance $\{\sigma_s^2\}_{s=1}^S$ adapted to the gradient norm, which may make the algorithm converge well. In practice, we do not have to do this and can instead inject homoscedastic noise while incorporating a carefully chosen scaling factor into the step size of the gradient descent. By doing so, the differential privacy of our FedDL with gradient perturbation can be guaranteed by Theorem \ref{thm:dp2}.

\subsection{Fed-tSNE+ and Fed-UMAP+}
Based on the above discussion, we propose the security-enhanced versions of Fed-tSNE and Fed-UMAP, denoted by Fed-tSNE+ and Fed-UMAP+, for which Algorithm \ref{alg_fedvisual} has noise injection in line 1 (Algorithm \ref{alg:feddl}).


\begin{table*}[!h]
\centering
\begin{tabular}{c|c||c|c}
\hline
        & Metric & k=5        & k=10       \\ \hline
\multirow{2}{*}{Fed-tSNE} & NMI  & 0.741$\pm$0.011     & 0.740$\pm$0.011 \\
& NPA 10-NN  & 0.336$\pm$0.001     & 0.338$\pm$0.001 \\
\hline
\multirow{2}{*}{Fed-tSNE+} & NMI  & 0.709$\pm$0.026     & 0.702$\pm$0.015 \\
& NPA 10-NN  & 0.286$\pm$0.001     & 0.283$\pm$0.001 \\
\hline
\end{tabular}
\caption{Performance (mean$\pm$std) of Fed-tSNE and Fed-tSNE+ on non-IID data for different values of the number of clients $k$}
\label{tab:nonIID-k}
\end{table*}

\begin{table*}[!h]
\centering
\begin{tabular}{c|c||c|c|c|c|c}
\hline
        & Metric & k=5        & k=10       & k=20       & k=50       & k=100      \\ \hline
\multirow{2}{*}{Fed-tSNE} & NMI  & 0.747$\pm$0.016     & 0.753$\pm$0.020 & 0.745$\pm$0.009 & 0.747$\pm$0.014 & 0.749$\pm$0.015 \\
& NPA 10-NN  & 0.337$\pm$0.001 & 0.337$\pm$0.001 & 0.337$\pm$0.001 & 0.338$\pm$0.001 & 0.338$\pm$0.001 \\
\hline
\multirow{2}{*}{Fed-tSNE+} & NMI  & 0.740$\pm$0.013 & 0.747$\pm$0.007 & 0.742$\pm$0.019 & 0.741$\pm$0.018 & 0.741$\pm$0.012 \\
& NPA 10-NN  & 0.323$\pm$0.001 & 0.326$\pm$0.001 & 0.329$\pm$0.001 & 0.332$\pm$0.001 & 0.333$\pm$0.001 \\
\hline
\end{tabular}
\caption{Performance (mean$\pm$std) of Fed-tSNE and Fed-tSNE+ on IID data for different values of the number of clients $k$}
\label{tab:IID-k}
\end{table*}

\begin{table*}[!h]
\centering
\begin{tabular}{c|c||c|c|c|c|c}
\hline
        & & & \multicolumn{2}{c|}{IID}  & \multicolumn{2}{c}{non-IID} \\ \hline \hline
& Metric & SpeClust & Fed-SpeClust & Fed-SpeClust+ & Fed-SpeClust & Fed-SpeClust+ \\ 
\hline
\multirow{2}{*}{MNIST}& NMI  & 0.5415$\pm$0.0009     & 0.5240$\pm$0.0038         & 0.5220$\pm$0.0052     & 0.5235$\pm$0.0051     & 0.5025$\pm$0.0068                \\
& ARI  & 0.3837$\pm$0.0008     & 0.3815$\pm$0.0076         & 0.3807$\pm$0.0088     & 0.3806$\pm$0.1123     & 0.3829$\pm$0.0102\\
\hline
\multirow{2}{*}{COIL-20}& NMI  & 0.8885$\pm$0.0016     & 0.8425$\pm$0.0218         & 0.8333$\pm$0.0173     & 0.8339$\pm$0.0216     & 0.8215$\pm$0.0163                \\
& ARI  & 0.6066$\pm$0.0012     & 0.5113$\pm$0.0557         & 0.4793$\pm$0.0426     & 0.4895$\pm$0.0639     & 0.4551$\pm$0.0322\\
\hline
\multirow{2}{*}{Mice-Protein}& NMI  & 0.3241$\pm$0.0063     & 0.3233$\pm$0.0121         & 0.3220$\pm$0.0143     & 0.3222$\pm$0.0190     & 0.3198$\pm$0.0100                \\
& ARI  & 0.1837$\pm$0.0037     & 0.1827$\pm$0.0033         & 0.1802$\pm$0.0154     & 0.1809$\pm$0.0024     & 0.1783$\pm$0.0016\\
 \hline
\end{tabular}

\caption{Performance (mean$\pm$std) of spectral clustering}
\label{SpeClust}
\end{table*}

\section{Experiments}
\subsection{Data Visualization}\label{sec_visualization}

We applied the proposed Fed-tSNE and Fed-UMAP methods to the MNIST and Fashion-MNIST datasets, with $m_X = 40,000$, and set $n_Y = 500$. We designed the experiment with 10 clients, where IID (independent and identically distributed) refers to each client’s data being randomly sampled from the MNIST dataset, thus including all classes. In contrast, non-IID means that each client’s data contains only a single class. After reducing the data dimension to two, we visualized them. Figure \ref{fig:MNIST} presents the results on MNIST, showing the data distribution under both IID and non-IID conditions. Additionally, we included results using Fed-tSNE+ and Fed-UMAP+, where noise $\boldsymbol{E}$ is introduced to the gradient $\nabla f_p(\boldsymbol{Y_p})$. Each element of $\boldsymbol{E}$ is drawn from $\mathcal{N}(0, \text{sd}^2(\nabla f_p(\boldsymbol{Y_p})))$, where $\text{sd}(\nabla f_p(\boldsymbol{Y_p}))$ represents the standard deviation of $\nabla f_p(\boldsymbol{Y_p})$. Due to space limitations, the results on Fashion-MNIST are provided in the Appendix (Figure \ref{fig:FASHION-MNIST}). Based on the visualization results, our proposed methods perform very well in all settings, with only minor differences compared to the non-distributed results. They preserved nearly all the essential information and structure of the data. Tables \ref{table-mnist} and \ref{table-fmnist} provide quantitative evaluations using the following metrics (detailed in Appendix A):
\textbf{CA (Classification Accuracy) } with k-NN, \textbf{NPA (Neighbor Preservation Accuracy) } with k-NN, \textbf{NMI (Normalized Mutual Information)} of k-means, and \textbf{SC (Silhouette Coefficient)} of k-means.
It can be observed that the performance of our proposed method shows a slight decline in various metrics compared to the nondistributed results, which is unavoidable. However, the overall differences remain within an acceptable range. Notably, the method performs slightly better on distributed data when the distribution is IID compared to non-IID. Moreover, the performance of Fed-tSNE+ and Fed-UMAP+ with added noise to protect privacy is somewhat inferior to the performance without noise, which is expected, as the non-IID scenario and the introduction of noise both impact the accuracy of $\bm Y$'s learning on whole $\bm X$, thereby affecting the final results. 

\begin{figure}[!h]
    \centering
    \includegraphics[width=0.45\textwidth]{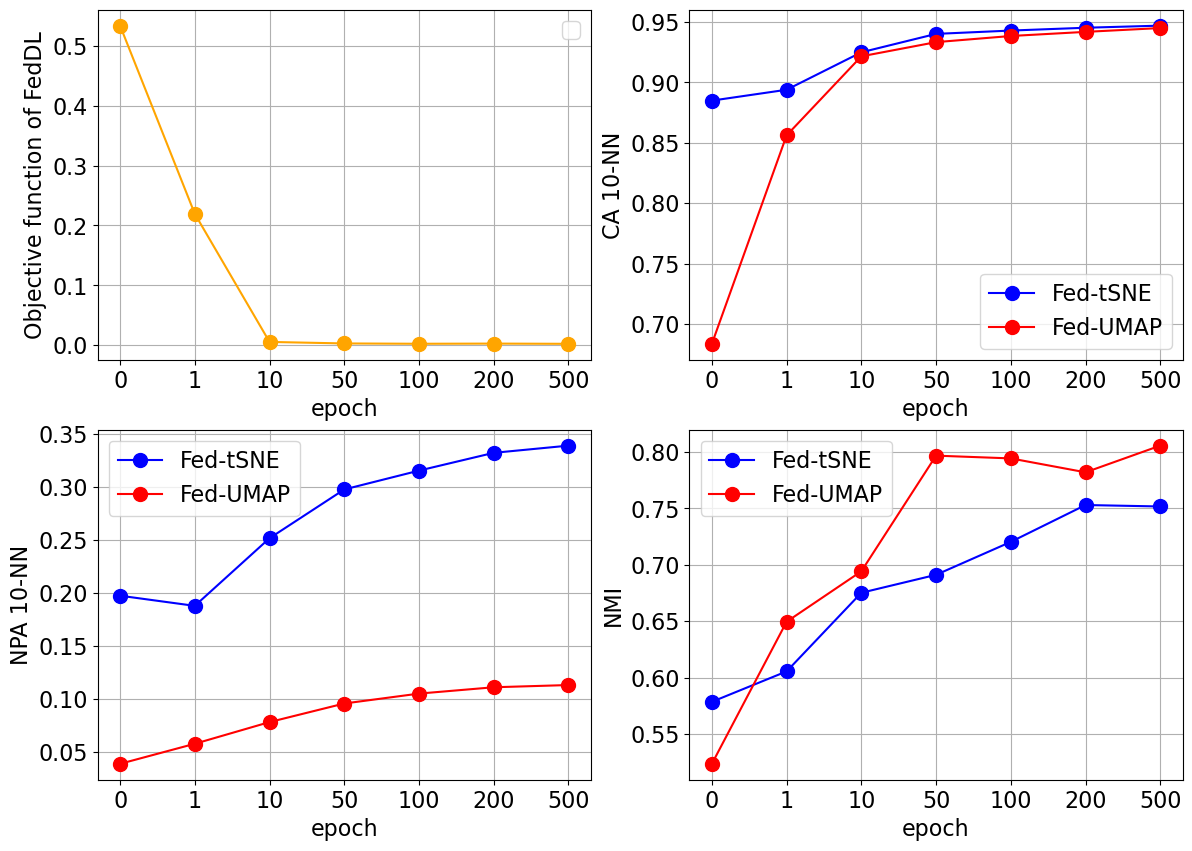}
    \caption{Convergence Performance on MNIST}
    \label{fig:convergence}
\end{figure}

\begin{figure*}[!h]
    \centering
    \includegraphics[width=0.7\textwidth, trim={10 5 0 5},clip]{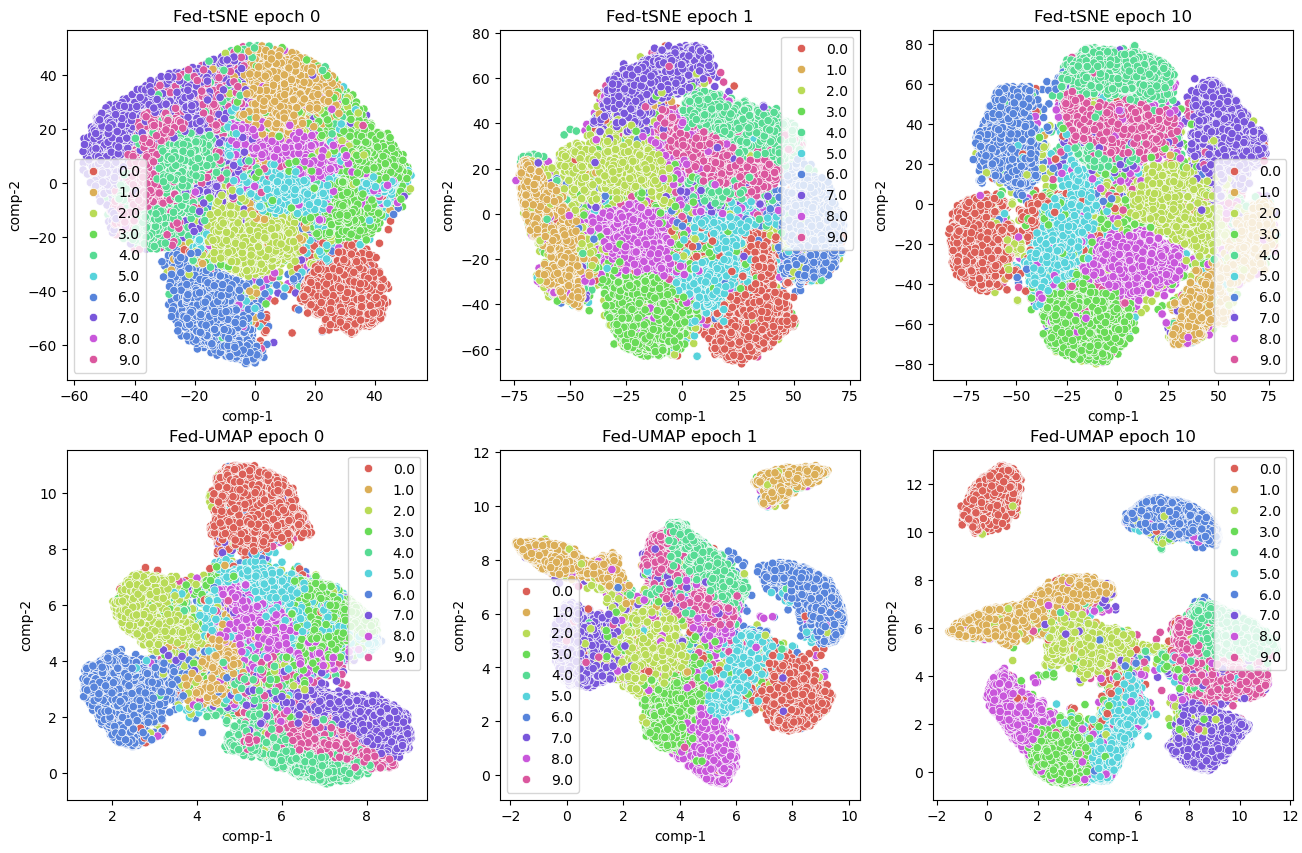}
    \caption{Visualization of Fed-tSNE and Fed-UMAP Convergence from epoch $1$ to $10$ (MNIST)}
    \label{fig:epoch_vis_0-10}
\end{figure*}

\paragraph{Convergence Analysis} We also conducted experiments to test the convergence of our methods. In Figure \ref{fig:convergence}, the relevant metrics reached convergence after approximately $50$ epochs. Figure \ref{fig:epoch_vis_0-10} provides a more intuitive demonstration that, with the increase in epochs, the learning of $\bm Y$ significantly improves the final results of Fed-tSNE and Fed-UMAP, further confirming the feasibility of our method. (The full process visualization is included in Figure \ref{fig:epoch_vis} of Appendix A.)

In addition, we also studied the impact of $n_y$ and noise level $\beta$ on NMI (Figures \ref{fig:ny} and \ref{fig:beta} in Appendix A). The noise level $\beta$ controls the scale of noise, with each element of noise $\boldsymbol{E}$ being drawn from $\mathcal{N}(0, \beta^2 \text{sd}^2(\nabla f_p(\boldsymbol{Y_p})))$. We see, regardless of the method or conditions, that the larger the $\bm Y$ volume or the smaller the noise level $\beta$ (indicating a lower privacy protection requirement), the better the NMI results.

Besides, in the current non-IID case, each client has one class of data, which is the hardest setting (the distribution of clients is highly heterogeneous), while the IID case is the easiest setting. Other settings interpolate between these two extreme cases. To further investigate the impact of the number of clients $k$, we conducted additional experiments on MNIST to explore the impact of client number. In the IID setting, the dataset is randomly divided among $k$ clients. In the non-IID setting, we focus on the extreme case of completely different distributions, adding $k = 5$, with each client containing data from two distinct classes. The results, shown in Table \ref{tab:nonIID-k} and \ref{tab:IID-k}, demonstrate that our proposed methods show stable performance across different values of $k$.

\subsection{Clustering performance}
We utilized three datasets MNIST, COIL-20, and Mice-Protein (detailed in Appendix) to evaluate the effectiveness of our Fed-SpeClust. The hyperparameters were adjusted accordingly and the corresponding results are presented in Table \ref{SpeClust}. In addition to the NMI metric used previously, we also employed the ARI (Adjusted Rand Index) metric, detailed in Appendix. 
We see that both NMI and ARI indicate that Fed-SpeClust achieves results comparable to the original spectral clustering, despite a slight decrease in performance, demonstrating the feasibility of our method.


\section{Conclusion}
This work proposed FedDL and applied it to t-SNE and UMAP to visualize distributed data. The idea was also extended for spectral clustering to cluster distributed data. We provided theoretical guarantees such as differential privacy. Experimental results demonstrated that the accuracies of our federated algorithms are close to the original algorithms. 

\section*{Acknowledgments}
This work was partially supported by the National Natural Science Foundation of China under Grant No.62376236 and the Guangdong Provincial Key Laboratory of Mathematical Foundations for Artificial Intelligence (2023B1212010001).


\bibliography{aaai25}

\clearpage

\bigskip
\appendix

\onecolumn
\section{More about The Experiments}

\subsection{Dataset Description}
The details about the datasets in the experiments are as follows.
\begin{itemize}
    \item \textbf{MNIST} \footnote{\url{http://yann.lecun.com/exdb/mnist/}} contains 70,000 images of handwritten digits (0-9), with 60,000 used for training and 10,000 for testing. Each image is 28x28 pixels.
    
    \item \textbf{Fashion-MNIST} \footnote{\url{https://github.com/zalandoresearch/fashion-mnist}} is a dataset of Zalando's article images consisting of 70,000 grayscale images in 10 categories, with 60,000 for training and 10,000 for testing. Each image is 28x28 pixels, designed to serve as a drop-in replacement for the original MNIST dataset. 
    
    \item \textbf{COIL-20 (The Columbia Object Image Library)} \footnote{\url{https://www.cs.columbia.edu/CAVE/software/softlib/coil-20.php}} contains 1,440 grayscale images of 20 objects. Each object was imaged at different angles, making the dataset useful for object recognition tasks. Each image is 32x32 pixels in size.
    
    \item \textbf{Mice-Protein (The Mice Protein Expression dataset)} \footnote{\url{https://archive.ics.uci.edu/dataset/342/mice+protein+expression}} consists of protein expression levels measured across 77 proteins for 72 mice.
\end{itemize}

\subsection{Detailed Definitions of Evaluation Metrics}
The details about the definition of the evaluation metrics CA, NPA, MNI, and SC are as follows.
\begin{itemize}
    \item \textbf{CA (Classification Accuracy) with k-NN} measures the classification accuracy of k-NN in the embedding space according to the true labels. In the experiments, we use $k = 1$, $10$, and $50$. The ratio between training data and testing data is $7:3$.

    \item \textbf{NPA (Neighbor Preservation Accuracy) with k-NN} measures the neighbor preservation accuracy of k-NN in the embedding space according to the true labels. Similar to CA, we use $k = 1$, $10$, and $50$.
    
    \item \textbf{NMI (Normalized Mutual Information)} measures the similarity between clustering results and true labels based on mutual information. NMI ranges from $0$ to $1$, with higher values indicating better clustering performance. In the experiments, we use k-means to attain clustering results in the embedding space.

    \item \textbf{SC (Silhouette Coefficient)} evaluates the quality of clustering by considering both intra-cluster cohesion and inter-cluster separation. The SC value ranges from $-1$ to $1$, where a value closer to $1$ indicates well-separated clusters. Similar to NMI, we use k-means to attain clustering results.
    \item \textbf{ARI (Adjusted Rand Index)} measures the similarity between predicted labels and true labels by analyzing how pairs of samples are assigned in both clustering and the value ranges from $-1$ to $1$, where $1$ indicates perfect agreement between the predicted and true labels.
\end{itemize}


\subsection{Additional Experimental Results}
\begin{figure*}[!h]
    \centering
    \includegraphics[width=1\textwidth]{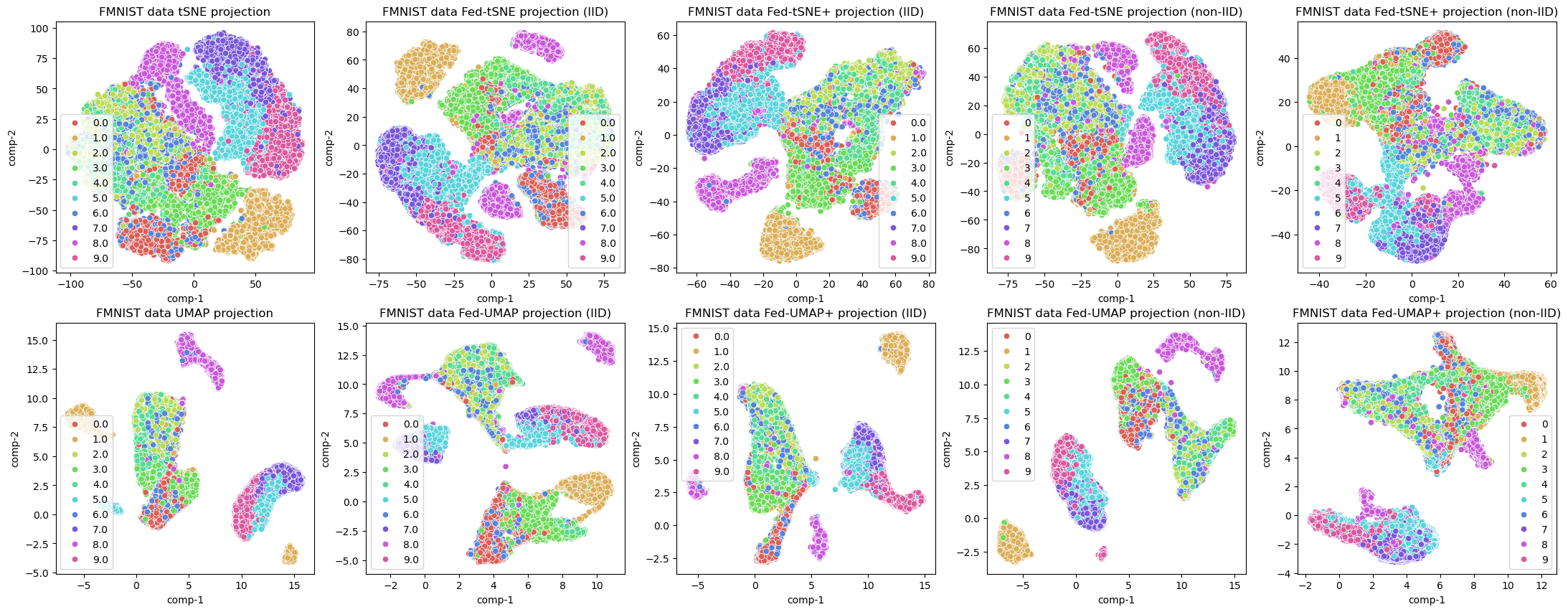}
    \caption{Visualization of Fashion-MNIST}
    \label{fig:FASHION-MNIST}
\end{figure*}

\begin{figure*}[!h]
    \centering
    \includegraphics[width=1\textwidth]{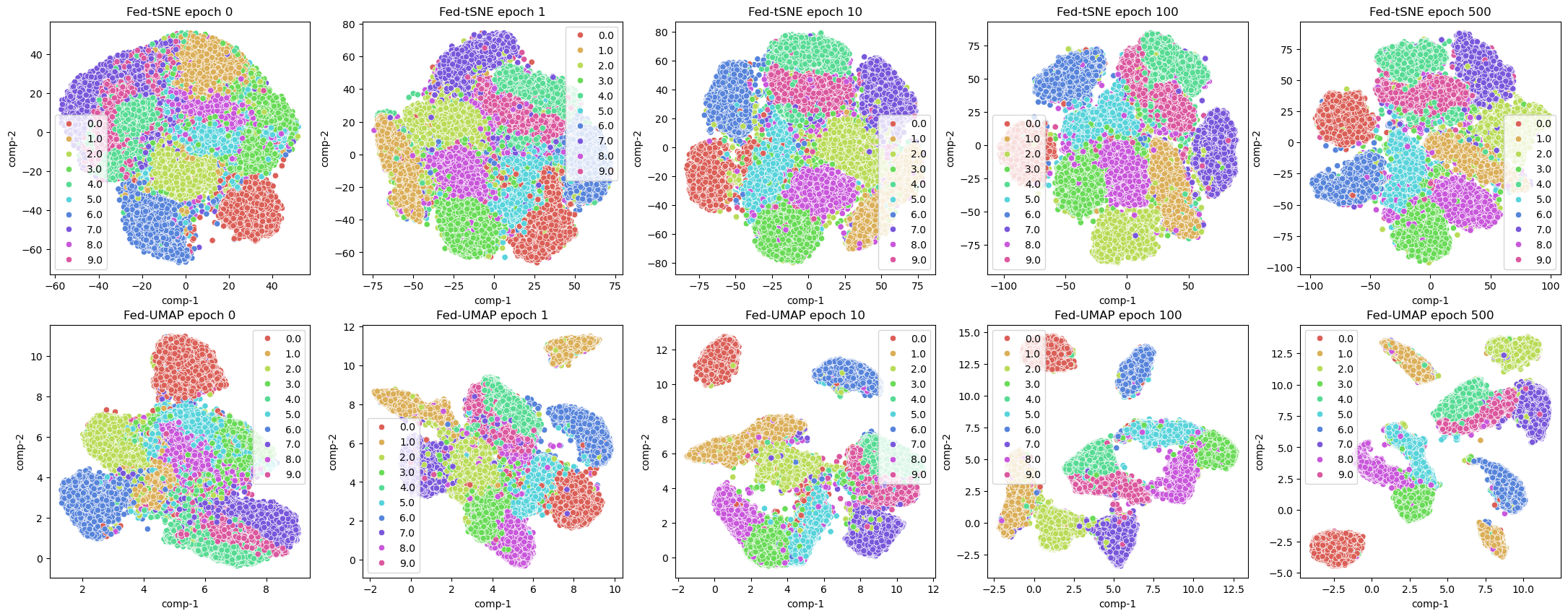}
    \caption{Visualization of Fed-tSNE and Fed-UMAP Convergence (MNIST)}
    \label{fig:epoch_vis}
\end{figure*}

\begin{figure}[!h]
    \centering
    \includegraphics[width=0.5\linewidth]{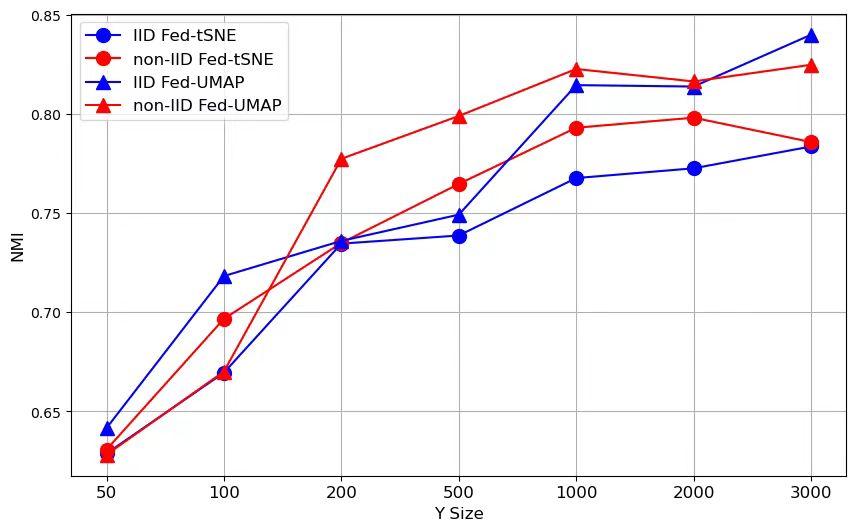}
    \caption{Impact of $n_y$ on NMI (MNIST)}
    \label{fig:ny}
\end{figure}

\begin{figure}[!h]
    \centering
    \includegraphics[width=0.5\linewidth]{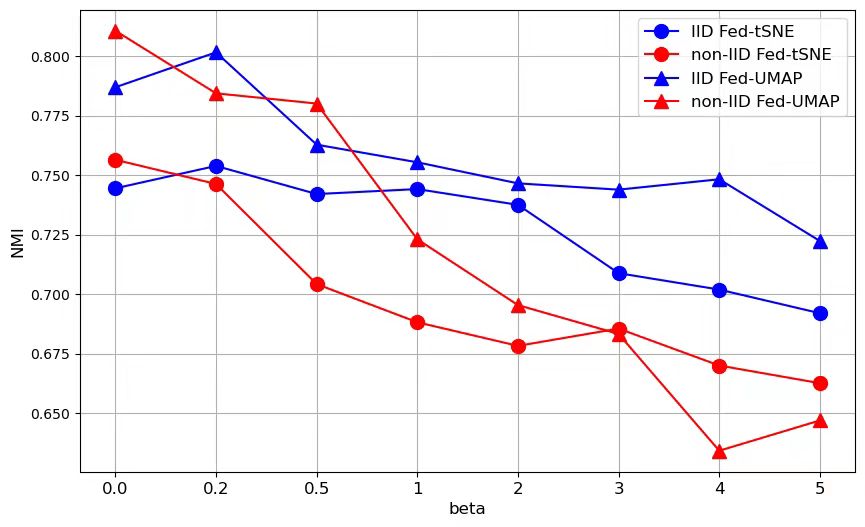}
    \caption{Impact of noise level $\beta$ on NMI (MNIST)}
    \label{fig:beta}
\end{figure}

We include the following experimental results:
\begin{itemize}
    \item \textbf{Figure \ref{fig:FASHION-MNIST}} presents the results on Fashion-MNIST, where the distribution follows either IID or non-IID patterns. Additionally, we included results using Fed-tSNE+ and Fed-UMAP+, where the variance of noise is the same as that of the gradients.

    \item \textbf{Figure \ref{fig:epoch_vis}} illustrates the convergence of Fed-tSNE and Fed-UMAP to their final results over 500 epochs on the MNIST dataset.
    
    \item \textbf{Figure \ref{fig:ny}} illustrates the effect of $n_y$ on the Normalized Mutual Information (NMI) for the MNIST dataset.
    
    \item \textbf{Figure \ref{fig:beta}} illustrates the effect of noise level $\beta$ on the Normalized Mutual Information (NMI) for the MNIST dataset. 
\end{itemize}

\section{Proof of Theorem \ref{thm:reconstruction error 1}}

\begin{lem}\label{lem:perturbation error 1}
Given $\bm X\in\mathbb R^{m\times n}$, let $\Tilde{\bm X}=\bm X+\bm\Delta$ be the perturbed data of $\bm X$ where all entries of $\bm\Delta=(e_{ij})_{i\in[m],j\in[n]}$ are sampled from $\mathcal{N}(0,\sigma^2)$. Then, with at least $1-n(n-1)e^{-t}$, the perturbation deviance on the Gaussian kernel matrix in terms of Frobenius norm can be bounded as
\begin{equation}
\begin{aligned}
&\left\Vert\bm{K}_{\Tilde{\bm X},\Tilde{\bm X}} - \bm{K}_{\bm X,\bm X}\right\Vert_F \le \sqrt{2}n\gamma\left[\sigma^2\xi_m^2 + 2\frac{\left\Vert D_{\bm X,\bm X}\right\Vert_\infty}{\sqrt{2}}\sigma\xi_m\right]
\end{aligned}
\end{equation}
where $\xi_m = \sqrt{m+\sqrt{2mt}+2t}$, $\gamma$ is the hyperparameter of the Gaussian kernel controlling the smoothness, and $\left\Vert D_{\bm X,\bm X}\right\Vert_\infty$ is the maximum entry of the pairwise euclidean distance on $\bm X$.
\end{lem}

\begin{proof}
We have the following derivation for the reconstruction of kernel matrix:
\begin{align*}
& \left\Vert\bm{K}_{\Tilde{\bm X},\Tilde{\bm X}} - \bm{K}_{\bm X,\bm X}\right\Vert_F^2\\
= & \sum_{i,j=1}^{n}\left(k(\Tilde{\bm x}_i,\Tilde{\bm x}_j) - k(\bm x_i,\bm x_j)\right)^2\\
= & \sum_{i,j=1}^{n}\left[k(\bm x_i + \bm e_i,\bm x_j + \bm e_j) - k(\bm x_i,\bm x_j)\right]^2\\
= & \sum_{i,j=1}^{n}\left[\exp\left(-\gamma\left\Vert\left(\bm x_i + \bm e_i\right) - \left(\bm x_j + \bm e_j\right)\right\Vert^2\right) -\exp\left(-\gamma\left\Vert\bm x_i - \bm x_j\right\Vert^2\right)\right]^2\\
\le & \sum_{i,j=1}^{n}\gamma^2\left[\left\Vert\left(\bm x_i + \bm e_i\right) - \left(\bm x_j + \bm e_j\right)\right\Vert^2 - \left\Vert\bm x_i - \bm x_j\right\Vert^2\right]^2\\
= & \sum_{i,j=1}^{n}\gamma^2\left[\left\Vert\left(\bm x_i - \bm x_j\right) + \left(\bm e_i - \bm e_j\right)\right\Vert^2 - \left\Vert\bm x_i - \bm x_j\right\Vert^2\right]^2\\
= & \sum_{i,j=1}^{n}\gamma^2\left[\left\Vert\bm e_i - \bm e_j\right\Vert^2 + 2\left\langle\bm x_i - \bm x_j,\bm e_i - \bm e_j\right\rangle\right]^2\\
\le & \sum_{i,j=1}^{n}\gamma^2\left[\left\Vert\bm e_i - \bm e_j\right\Vert^2 + 2\left\Vert\bm x_i - \bm x_j\right\Vert\left\Vert\bm e_i - \bm e_j\right\Vert\right]^2
\end{align*}
where the first inequality follows from the fact that the exponential function is locally Lipschitz continuous, \textit{i.e.}, $|e^x - e^y| \le |x - y|$ for $x,y\le 0$. Note that $\frac{1}{2\sigma^2}\left\Vert\bm e_i - \bm e_j\right\Vert^2 = \sum_{k = 1}^m(\frac{e_{k,i} - e_{k,j}}{\sqrt{2}\sigma})^2\sim \chi_m^2$. By \cite{laurent2000adaptive}, one can give the bound with probability at least $1 - e^{-t}$
\begin{equation}
\begin{aligned}
\frac{1}{2\sigma^2}\left\Vert\bm e_i - \bm e_j\right\Vert^2 \le m + 2\sqrt{mt} + 2t
\end{aligned}
\end{equation}
which implies the the union bound 
\begin{equation}
\begin{aligned}
\max_{i,j}\left\Vert\bm e_i - \bm e_j\right\Vert^2 \le 2\sigma^2(m + 2\sqrt{mt} + 2t)
\end{aligned}
\end{equation}
with probability at least $1-n(n-1)e^{-t}$. Since $\forall i\in[n],j\in[n],\left\Vert\bm x_i - \bm x_j\right\Vert_2\le\left\Vert \bm{D}_{\bm X,\bm X}\right\Vert_\infty$ where $\bm{D}_{\bm X,\bm X}$ is the pairwise euclidean distance between $\bm X$ and $\bm X$, and let $\xi_m = \sqrt{m + 2\sqrt{mt} + 2t}$, it follows that with probability at least $1 - n(n-1)e^{-t}$
\begin{align*}
& \left\Vert\bm{K}_{\Tilde{\bm X},\Tilde{\bm X}} - \bm{K}_{\bm X,\bm X}\right\Vert_F^2 \le \sum_{i,j=1}^{n}\gamma^2\left[2\sigma^2\xi_m^2 + 2\left\Vert D_{\bm X,\bm X}\right\Vert_\infty\sqrt{2}\sigma\xi_m\right]^2 = 2n^2\gamma^2\left[\sigma^2\xi_m^2 + 2\frac{\left\Vert D_{\bm X,\bm X}\right\Vert_\infty}{\sqrt{2}}\sigma\xi_m\right]^2
\end{align*}

\end{proof}

\begin{proof}[Proof of Theorem \ref{thm:reconstruction error 1}]
Denote by $\bm X_a$ the augmented matrix $[\bm Y, \bm X] = [\bm y_1,\dots,\bm y_{n_y},\bm x_1,\dots,\bm x_{n_x}]$ and let $\Tilde{\bm X}_a^x = [\bm Y, \Tilde{\bm X}]$ be the noisy augmented matrix. Let $\bm C = \bm K_{\Tilde{\bm X}_a^x,\bm Y}$, $\bm W = \bm K_{\bm Y,\bm Y}$ with $\bm W_k^\dagger$ being the best rank-$k$ approximation in terms of the spectral norm. For convenience, we omit the $\bm{Y}$ in $\widehat{\bm H}_{\Tilde{\bm X},\Tilde{\bm X}|\bm{Y}}$. It follows from the triangular inequality of matrix norm that 
\begin{align*}
&\Vert\widehat{\bm H}_{\Tilde{\bm X},\Tilde{\bm X}} - \bm{K}_{\bm X,\bm X}\Vert_2\\
= & \Vert\widehat{\bm H}_{\Tilde{\bm X},\Tilde{\bm X}} - \bm{K}_{\Tilde{\bm X},\Tilde{\bm X}} + \bm{K}_{\Tilde{\bm X},\Tilde{\bm X}} - \bm{K}_{\bm X,\bm X}\Vert_2\\
\le & \Vert\widehat{\bm H}_{\Tilde{\bm X},\Tilde{\bm X}} - \bm{K}_{\Tilde{\bm X},\Tilde{\bm X}}\Vert_2 + \Vert\bm{K}_{\Tilde{\bm X},\Tilde{\bm X}} - \bm{K}_{\bm X,\bm X}\Vert_2\\
= & \underbrace{\Vert\bm C\bm W_k^\dagger\bm C^T - \bm{K}_{\Tilde{\bm X}_a^x,\Tilde{\bm X}_a^x}\Vert_2}_{T_1} + \underbrace{\Vert\bm{K}_{\Tilde{\bm X},\Tilde{\bm X}} - \bm{K}_{\bm X,\bm X}\Vert_2}_{T_2}
\end{align*}
where the perturbation term $T_2$ depends on both data $\bm X$ and the noise intensity and is irrelevant to our federated optimization process.

By Lemma \ref{lem:perturbation error 1}, we have an upper bound on $T_2$.
\begin{align*}
&\Vert\bm{K}_{\Tilde{\bm X},\Tilde{\bm X}} - \bm{K}_{\bm X,\bm X}\Vert_2 \le \Vert\bm{K}_{\Tilde{\bm X},\Tilde{\bm X}} - \bm{K}_{\bm X,\bm X}\Vert_F \le \sqrt{2}n_x\gamma\left[\sigma^2\xi_m^2 + 2\frac{\left\Vert D_{\bm X,\bm X}\right\Vert_\infty}{\sqrt{2}}\sigma\xi_m\right]
\end{align*}

We will show that the approximation term $T_1$ depends on both Nystr\"om approximation mechanism and our proposed FedMMD through $\Tilde{\bm X},\bm Y$. 

Since $\bm{K}_{\Tilde{\bm X}_a^x,\Tilde{\bm X}_a^x}$ is positive semi-definite, we assume that $\bm{K}_{\Tilde{\bm X}_a^x,\Tilde{\bm X}_a^x} = \bm A^T\bm A$ for some matrix $\bm A = [\bm a_1,\cdots,\bm a_{n_y},\bm a_{n_y + 1},\cdots,\bm a_{n_x + n_y}]\in\mathbb R^{\ell\times (n_x + n_y)}$ with $\ell\ge k$. $\Bar{\bm{K}}_{\Tilde{\bm X}_a^x,\Tilde{\bm X}_a^x}$ is the best rank-$k$ approximation of $\bm{K}_{\Tilde{\bm X}_a^x,\Tilde{\bm X}_a^x}$. Let $\bm S\in\{0,1\}^{n\times n}$ be the selection matrix such that $\bm C = \bm{K}_{\Tilde{\bm X}_a^x,\bm Y} = \bm{K}_{\Tilde{\bm X}_a^x,\Tilde{\bm X}_a^x}\bm S$. Then, denoting $\bm C_{\bm A} = \bm A\bm S = [\bm a_1,\cdots,\bm a_{n_y}]$, we have by Theorem 3 in \cite{JMLR:v6:drineas05a}
\begin{align*}
&\Vert\bm C\bm W_k^\dagger\bm C^T - \bm{K}_{\Tilde{\bm X}_a^x,\Tilde{\bm X}_a^x}\Vert_2\\
\le & \Vert\bm{K}_{\Tilde{\bm X}_a^x,\Tilde{\bm X}_a^x} - \Bar{\bm{K}}_{\Tilde{\bm X}_a^x,\Tilde{\bm X}_a^x}\Vert_2 + 2\Vert \bm A\bm A^T - \bm C_{\bm A}\bm C_{\bm A}^T\Vert_2\\
= & \Vert\bm{K}_{\Tilde{\bm X}_a^x,\Tilde{\bm X}_a^x} - \Bar{\bm{K}}_{\Tilde{\bm X}_a^x,\Tilde{\bm X}_a^x}\Vert_2 + 2\left\Vert\sum_{i = 1}^{n_x + n_y} \bm a_i\bm a_i^T - \sum_{i = 1}^{n_y} \bm a_i\bm a_i^T\right\Vert_2\\
= & \Vert\bm{K}_{\Tilde{\bm X}_a^x,\Tilde{\bm X}_a^x} - \Bar{\bm{K}}_{\Tilde{\bm X}_a^x,\Tilde{\bm X}_a^x}\Vert_2 + 2\Vert\sum_{i = 1}^{n_x}\bm a_{i - n_y}\bm a_{i - {n_y}}^T\Vert_2\\
\le & \Vert\bm{K}_{\Tilde{\bm X}_a^x,\Tilde{\bm X}_a^x} - \Bar{\bm{K}}_{\Tilde{\bm X}_a^x,\Tilde{\bm X}_a^x}\Vert_2 + 2\Vert \bm A_{:,(n_y + 1):}\bm A_{:,(n_y + 1):}^T\Vert_2\\
\le & \Vert\bm{K}_{\Tilde{\bm X}_a^x,\Tilde{\bm X}_a^x} - \Bar{\bm{K}}_{\Tilde{\bm X}_a^x,\Tilde{\bm X}_a^x}\Vert_2 + 2\Vert \bm A_{:,(n_y + 1):}^T\bm A_{:,(n_y + 1):}\Vert_2\\
\le & \Vert\bm{K}_{\Tilde{\bm X}_a^x,\Tilde{\bm X}_a^x} - \Bar{\bm{K}}_{\Tilde{\bm X}_a^x,\Tilde{\bm X}_a^x}\Vert_2 + 2\Vert\bm K_{\Tilde{\bm X},\Tilde{\bm X}}\Vert_2\\
\le & \Vert\bm{K}_{\Tilde{\bm X}_a^x,\Tilde{\bm X}_a^x} - \Bar{\bm{K}}_{\Tilde{\bm X}_a^x,\Tilde{\bm X}_a^x}\Vert_2 + 2n_x\\
= & \sigma_{k + 1}(\bm{K}_{\Tilde{\bm X}_a^x,\Tilde{\bm X}_a^x}) + 2n_x\\
\le & \frac{\sigma_1(\bm{K}_{\Tilde{\bm X}_a^x,\Tilde{\bm X}_a^x})}{\sigma_{n_x + n_y}(\bm{K}_{\Tilde{\bm X}_a^x,\Tilde{\bm X}_a^x})}\sigma_{n_x + n_y}(\bm{K}_{\Tilde{\bm X}_a^x,\Tilde{\bm X}_a^x}) + 2n_x\\
\le & \frac{\sigma_1(\bm{K}_{\Tilde{\bm X}_a^x,\Tilde{\bm X}_a^x})}{\sigma_{n_x + n_y}(\bm{K}_{\Tilde{\bm X}_a^x,\Tilde{\bm X}_a^x})}\frac{\mid\bm I_*^T\bm{K}_{\Tilde{\bm X}_a^x,\Tilde{\bm X}_a^x}\bm I_*\mid}{\bm I_*^T\bm I_*} + 2n_x\\
= & \textup{Cond}(\bm{K}_{\Tilde{\bm X}_a^x,\Tilde{\bm X}_a^x})\frac{\mid\textup{MMD}(\Tilde{\bm X},\bm Y) + (n_x + n_y)\mid}{n_x + n_y} + 2n_x\\
\le & \textup{Cond}(\bm{K}_{\Tilde{\bm X}_a^x,\Tilde{\bm X}_a^x})\left(\frac{\mid\textup{MMD}(\Tilde{\bm X},\bm Y)\mid}{n_x + n_y} + 1\right) + 2n_x
\end{align*}
where $\bm I_* = [-\bm I_{n_y}^T,\bm I_{n_x}^T]^T$ for $\bm I_{n_y}\in\{1\}^{n_y},\bm I_{n_x}\in\{1\}^{n_x}$, and we used the fact for the third inequality that $\|\bm A\bm A^T\|_2 = \|\bm A^T\bm A\|_2 \le \textup{Trace}(\bm A^T\bm A)$; used the variational characteristics of singular value decomposition for the penultimate inequality that $\sigma_n(\bm H)\le \frac{\bm x^T\bm H\bm y}{\|\bm x\|_2\|\bm y\|_2}\le\sigma_1(\bm H)$ for any matrix $K\in\mathbb R^{m\times n}$ and vectors $\bm x\in\mathbb R^m$, $\bm y\in\mathbb R^n$.

Combining upper bounds on $T_1$ and $T_2$, we have
\begin{align*}
&\Vert\widehat{\bm H}_{\Tilde{\bm X},\Tilde{\bm X}} - \bm{K}_{\bm X,\bm X}\Vert_F \le \textup{Cond}(\bm{K}_{\Tilde{\bm X}_a^x,\Tilde{\bm X}_a^x})\left(\frac{\mid\textup{MMD}(\Tilde{\bm X},\bm Y)\mid}{n_x + n_y} + 1\right) + 2n_x + \sqrt{2}n_x\gamma\left[\sigma^2\xi_m^2 + 2\frac{\left\Vert D_{\bm X,\bm X}\right\Vert_\infty}{\sqrt{2}}\sigma\xi_m\right]
\end{align*}

Analogously, we derive an upper bound on the approximation error in terms of Frobenius norm.
\begin{align*}
&\Vert\widehat{\bm H}_{\Tilde{\bm X},\Tilde{\bm X}} - \bm{K}_{\bm X,\bm X}\Vert_F\\
= & \Vert\widehat{\bm H}_{\Tilde{\bm X},\Tilde{\bm X}} - \bm{K}_{\Tilde{\bm X},\Tilde{\bm X}} + \bm{K}_{\Tilde{\bm X},\Tilde{\bm X}} - \bm{K}_{\bm X,\bm X}\Vert_F\\
\le & \Vert\widehat{\bm H}_{\Tilde{\bm X},\Tilde{\bm X}} - \bm{K}_{\Tilde{\bm X},\Tilde{\bm X}}\Vert_F + \Vert\bm{K}_{\Tilde{\bm X},\Tilde{\bm X}} - \bm{K}_{\bm X,\bm X}\Vert_F\\
= & \underbrace{\Vert\bm C\bm W_k^\dagger\bm C^T - \bm{K}_{\Tilde{\bm X}_a^x,\Tilde{\bm X}_a^x}\Vert_F}_{T_3} + \underbrace{\Vert\bm{K}_{\Tilde{\bm X},\Tilde{\bm X}} - \bm{K}_{\bm X,\bm X}\Vert_F}_{T_4}
\end{align*}

By Lemma \ref{lem:perturbation error 1}, we have an upper bound on $T_4$.
\begin{align*}
&\Vert\bm{K}_{\Tilde{\bm X},\Tilde{\bm X}} - \bm{K}_{\bm X,\bm X}\Vert_F \le \sqrt{2}n_x\gamma\left[\sigma^2\xi_m^2 + 2\frac{\left\Vert D_{\bm X,\bm X}\right\Vert_\infty}{\sqrt{2}}\sigma\xi_m\right]
\end{align*}

We will show that the approximation error $T_3$ depends on both Nystr\"om approximation mechanism and our proposed FedMMD through $\Tilde{\bm X},\bm Y$.

For $T_1$, 
\begin{align*}
&\Vert\bm C\bm W_k^\dagger\bm C^T - \bm{K}_{\Tilde{\bm X}_a^x,\Tilde{\bm X}_a^x}\Vert_F^2\\
\le & \Vert\bm{K}_{\Tilde{\bm X}_a^x,\Tilde{\bm X}_a^x} - \Bar{\bm{K}}_{\Tilde{\bm X}_a^x,\Tilde{\bm X}_a^x}\Vert_F^2 + 4\sqrt{k}\left\Vert \bm A\bm A^T\bm A\bm A^T - \bm C_{\bm A}\bm C_{\bm A}^T\bm C_{\bm A}\bm C_{\bm A}^T\right\Vert_F\\
= & \Vert\bm{K}_{\Tilde{\bm X}_a^x,\Tilde{\bm X}_a^x} - \Bar{\bm{K}}_{\Tilde{\bm X}_a^x,\Tilde{\bm X}_a^x}\Vert_F^2 + 4\sqrt{k}\left\Vert \bm A\bm A^T\bm A\bm A^T -  \bm A\bm A^T \bm C_{\bm A}\bm C_{\bm A}^T +  \bm A\bm A^T \bm C_{\bm A}\bm C_{\bm A}^T -  \bm C_{\bm A}\bm C_{\bm A}^T\bm C_{\bm A}\bm C_{\bm A}^T\right\Vert_F\\
= & \Vert\bm{K}_{\Tilde{\bm X}_a^x,\Tilde{\bm X}_a^x} - \Bar{\bm{K}}_{\Tilde{\bm X}_a^x,\Tilde{\bm X}_a^x}\Vert_F^2 + 4\sqrt{k}\left\Vert \bm A\bm A^T\left[ \bm A\bm A^T-  \bm C_{\bm A}\bm C_{\bm A}^T\right] + \left[ \bm A\bm A^T -  \bm C_{\bm A}\bm C_{\bm A}^T\right] \bm C_{\bm A}\bm C_{\bm A}^T\right\Vert_F\\
\le & \Vert\bm{K}_{\Tilde{\bm X}_a^x,\Tilde{\bm X}_a^x} - \Bar{\bm{K}}_{\Tilde{\bm X}_a^x,\Tilde{\bm X}_a^x}\Vert_F^2 + 4\sqrt{k}\left\{\left\Vert \bm A\bm A^T\right\Vert_F\left\Vert \bm A\bm A^T-  \bm C_{\bm A}\bm C_{\bm A}^T\right\Vert_2 + \left\Vert \bm A\bm A^T -  \bm C_{\bm A}\bm C_{\bm A}^T\right\Vert_2\left\Vert \bm C_{\bm A}\bm C_{\bm A}^T\right\Vert_F\right\}\\
= & \Vert\bm{K}_{\Tilde{\bm X}_a^x,\Tilde{\bm X}_a^x} - \Bar{\bm{K}}_{\Tilde{\bm X}_a^x,\Tilde{\bm X}_a^x}\Vert_F^2 + 4\sqrt{k}\left(\left\Vert \bm A\bm A^T\right\Vert_F + \left\Vert \bm C_{\bm A}\bm C_{\bm A}^T\right\Vert_F\right)\left\Vert \bm A\bm A^T-  \bm C_{\bm A}\bm C_{\bm A}^T\right\Vert_2\\
= & \Vert\bm{K}_{\Tilde{\bm X}_a^x,\Tilde{\bm X}_a^x} - \Bar{\bm{K}}_{\Tilde{\bm X}_a^x,\Tilde{\bm X}_a^x}\Vert_F^2 + 4\sqrt{k}\left(\left\Vert \bm K_{\Tilde{\bm X}_a^x,\Tilde{\bm X}_a^x}\right\Vert_F + \left\Vert \bm K_{\bm Y,\bm Y}\right\Vert_F\right)\left\Vert \bm A\bm A^T-  \bm C_{\bm A}\bm C_{\bm A}^T\right\Vert_2\\
\le & \Vert\bm{K}_{\Tilde{\bm X}_a^x,\Tilde{\bm X}_a^x} - \Bar{\bm{K}}_{\Tilde{\bm X}_a^x,\Tilde{\bm X}_a^x}\Vert_F^2 + 4\sqrt{k}\left(n_x + 2n_y\right)n_x
\end{align*}
where we used the fact for the last inequality that $\|\bm H\|_F = \sqrt{\sum_{i,j}k_{i,j}^2} \le \sqrt{\sum_{i,j}1} = n$ for a Gaussian kernel matrix $\bm H\in\mathbb R^{n\times n}$.

Then, it follows from the fact that $\forall a,b\in\mathbb R_+, (a^2 + b^2)^{1/2}\le a + b$ that
\begin{align*}
&\Vert\bm C\bm W_k^\dagger\bm C^T - \bm{K}_{\Tilde{\bm X}_a^x,\Tilde{\bm X}_a^x}\Vert_F\\
\le & \Vert\bm{K}_{\Tilde{\bm X}_a^x,\Tilde{\bm X}_a^x} - \Bar{\bm{K}}_{\Tilde{\bm X}_a^x,\Tilde{\bm X}_a^x}\Vert_F + 2k^{1/4}\sqrt{\left(n_x + 2n_y\right)n_x}\\
= & \sqrt{\sum_{i>k}\sigma_i^2(\bm{K}_{\Tilde{\bm X}_a^x,\Tilde{\bm X}_a^x})} + 2k^{1/4}n_x\sqrt{\left(1 + \frac{n_y}{n_x}\right)}\\
\le & \sqrt{(n_x + n_y - k)\sigma_1^2(\bm{K}_{\Tilde{\bm X}_a^x,\Tilde{\bm X}_a^x})} + 2k^{1/4}n_x\sqrt{\left(1 + \frac{n_y}{n_x}\right)}\\
= & \sqrt{(n_x + n_y - k)\frac{\sigma_1^2(\bm{K}_{\Tilde{\bm X}_a^x,\Tilde{\bm X}_a^x})}{\sigma_{n_x + n_y}^2(\bm{K}_{\Tilde{\bm X}_a^x,\Tilde{\bm X}_a^x})}\sigma_{n_x + n_y}^2(\bm{K}_{\Tilde{\bm X}_a^x,\Tilde{\bm X}_a^x})} + 2k^{1/4}n_x\sqrt{\left(1 + \frac{n_y}{n_x}\right)}\\
\le & \sqrt{n_x + n_y - k}\textup{Cond}(\bm{K}_{\Tilde{\bm X}_a^x,\Tilde{\bm X}_a^x})\frac{\mid\bm I_*^T\bm{K}_{\Tilde{\bm X}_a^x,\Tilde{\bm X}_a^x}\bm I_*\mid}{\bm I_*^T\bm I_*} + 2k^{1/4}n_x\sqrt{\left(1 + \frac{n_y}{n_x}\right)}\\
\le & \sqrt{n_x + n_y - k}\textup{Cond}(\bm{K}_{\Tilde{\bm X}_a^x,\Tilde{\bm X}_a^x})\left(\frac{\mid\textup{MMD}(\Tilde{\bm X},\bm Y)\mid}{n_x + n_y} + 1\right) + 2k^{1/4}n_x\sqrt{\left(1 + \frac{n_y}{n_x}\right)}
\end{align*}

Combining upper bounds on $T_3$ and $T_4$, we have
\begin{align*}
\Vert\widehat{\bm H}_{\Tilde{\bm X},\Tilde{\bm X}} - \bm{K}_{\bm X,\bm X}\Vert_F \le & \sqrt{n_x + n_y - k}\textup{Cond}(\bm{K}_{\Tilde{\bm X}_a^x,\Tilde{\bm X}_a^x})\left(\frac{\mid\textup{MMD}(\Tilde{\bm X},\bm Y)\mid}{n_x + n_y} + 1\right) + 2k^{1/4}n_x\sqrt{\left(1 + \frac{n_y}{n_x}\right)}\\
& + \sqrt{2}n_x\gamma\left[\sigma^2\xi_m^2 + 2\frac{\left\Vert D_{\bm X,\bm X}\right\Vert_\infty}{\sqrt{2}}\sigma\xi_m\right]
\end{align*}

\end{proof}

\section{Proof of Theorem \ref{thm:dp1}}
\begin{proof}
As declared by Definition 3.8 in \cite{dwork2014algorithmic}, the $\ell_2-$sensitivity of a function $f:\mathbb N^{|\mathcal{X}|}\to\mathbb R^k$ is
\begin{equation}
\begin{aligned}
\triangle^2(f) = \sup_{x\sim y}\Vert f(\bm x) - f(\bm y)\Vert_2
\end{aligned}
\end{equation}
where $\bm x\sim \bm y$ denotes that $\bm x$ and $\bm y$ are neighboring datasets. In our case, $f(\bm x) = \bm x$. Thus, the $\ell^2-$sensitivity of $f(\bm x) = \bm x$ is
\begin{equation}
\begin{aligned}
\triangle^2(f) = \sup_{\bm x\sim \bm y}\Vert f(\bm x) - f(\bm y)\Vert_2 \le \sup_{\bm x\sim \bm y}\Vert \bm x - \bm y\Vert_2 \le 2\tau_X
\end{aligned}
\end{equation}
Therefore, by Theorem 3.22 and Proposition 2.1 (Post-Processing) in \cite{dwork2014algorithmic}, our proposed Algorithm \ref{alg:feddl} is $(\varepsilon,\delta)-$differentially private if $\delta\ge 2c\tau_X/\varepsilon$, where $c^2 > 2\ln(1.25/\delta)$.

\end{proof}

\section{Proof of Theorem \ref{thm:reconstruction error 2}}
\begin{proof}
Denote by $\bm X_a$ the augmented matrix $[\bm Y, \bm X] = [\bm y_1,\dots,\bm y_\ell,\bm x_1,\dots,\bm y_n]$ and let $\Tilde{\bm X}_a^y = [\Tilde{\bm Y}, \bm X]$ be the noisy augmented matrix. Let $\Tilde{\bm C} = \bm K_{\Tilde{\bm X}_a^y,\Tilde{\bm Y}}$, $\Tilde{\bm W} = \bm K_{\Tilde{\bm Y},\Tilde{\bm Y}}$ with $\Tilde{\bm W}_k^\dagger$ being the Moore-Penrose inverse of the best rank-$k$ approximation in terms of the spectral norm. Since $\bm{K}_{\Tilde{\bm X}_a^y,\Tilde{\bm X}_a^y}$ is positive semi-definite, we assume that $\bm{K}_{\Tilde{\bm X}_a^y,\Tilde{\bm X}_a^y} = \bm A^T\bm A$ for some matrix $\bm A = [\bm a_1,\cdots,\bm a_{n_y},\bm a_{n_y + 1},\cdots,\bm a_{n_x + n_y}]\in\mathbb R^{\ell\times (n_x + n_y)}$ with $\ell\ge k$. Let $\bm S\in\{0,1\}^{n\times n}$ be the selection matrix such that $\Tilde{\bm C} = \bm{K}_{\Tilde{\bm X}_a^y,\Tilde{\bm Y}} = \bm{K}_{\Tilde{\bm X}_a^y,\Tilde{\bm X}_a^y}\bm S$. Then, denoting $\bm C_{\bm A} = \bm A\bm S = [\bm a_1,\cdots,\bm a_{n_y}]$, we have by Theorem 3 in \cite{JMLR:v6:drineas05a}
\begin{align*}
&\left\Vert\widehat{\bm H}_{\bm X, \bm{X}|\Tilde{\bm Y}} - \bm{K}_{\bm X, \bm X}\right\Vert_2\\
= & \left\Vert\Tilde{\bm C}\Tilde{\bm W}_k^\dagger\Tilde{\bm C}^T - \bm{K}_{\Tilde{\bm X}_a^y, \Tilde{\bm X}_a^y}\right\Vert_2\\
\le & \left\Vert\bm{K}_{\Tilde{\bm X}_a^y, \Tilde{\bm X}_a^y} - \Bar{\bm{K}}_{\Tilde{\bm X}_a^y,\Tilde{\bm X}_a^y}\right\Vert_2 + 2\left\Vert \bm A\bm A^T - \bm C_{\bm A}\bm C_{\bm A}\right\Vert_2\\
= & \left\Vert\bm{K}_{\Tilde{\bm X}_a^y, \Tilde{\bm X}_a^y} - \Bar{\bm{K}}_{\Tilde{\bm X}_a^y,\Tilde{\bm X}_a^y}\right\Vert_2 + 2\left\Vert \bm A_{:,(n_y + 1):}\bm A_{:,(n_y + 1):}^T\right\Vert_2\\
= & \left\Vert\bm{K}_{\Tilde{\bm X}_a^y, \Tilde{\bm X}_a^y} - \Bar{\bm{K}}_{\Tilde{\bm X}_a^y,\Tilde{\bm X}_a^y}\right\Vert_2 + 2\left\Vert \bm A_{:,(n_y + 1):}^T\bm A_{:,(n_y + 1):}\right\Vert_2\\
= & \left\Vert\bm{K}_{\Tilde{\bm X}_a^y, \Tilde{\bm X}_a^y} - \Bar{\bm{K}}_{\Tilde{\bm X}_a^y,\Tilde{\bm X}_a^y}\right\Vert_2 + 2\left\Vert \bm K_{\bm X,\bm X}\right\Vert_2\\
\le & \left\Vert\bm{K}_{\Tilde{\bm X}_a^y, \Tilde{\bm X}_a^y} - \Bar{\bm{K}}_{\Tilde{\bm X}_a^y,\Tilde{\bm X}_a^y}\right\Vert_2 + 2n_x\\
= & \sigma_{k + 1}\left(\bm{K}_{\Tilde{\bm X}_a^y, \Tilde{\bm X}_a^y}\right) + 2n_x\\
\le & \sigma_1\left(\bm{K}_{\Tilde{\bm X}_a^y, \Tilde{\bm X}_a^y}\right) + 2n_x\\
\le & \frac{\sigma_1\left(\bm{K}_{\Tilde{\bm X}_a^y, \Tilde{\bm X}_a^y}\right)}{\sigma_{n_x + n_y}\left(\bm{K}_{\Tilde{\bm X}_a^y, \Tilde{\bm X}_a^y}\right)}\sigma_{n_x + n_y}\left(\bm{K}_{\Tilde{\bm X}_a^y, \Tilde{\bm X}_a^y}\right) + 2n_x\\
\le & \frac{\sigma_1(\bm{K}_{\Tilde{\bm X}_a^y,\Tilde{\bm X}_a^y})}{\sigma_{n_x + n_y}(\bm{K}_{\Tilde{\bm X}_a^y,\Tilde{\bm X}_a^y})}\frac{\mid\bm I_*^T\bm{K}_{\Tilde{\bm X}_a^y,\Tilde{\bm X}_a^y}\bm I_*\mid}{\bm I_*^T\bm I_*} + 2n_x\\
= & \textup{Cond}(\bm{K}_{\Tilde{\bm X}_a^y,\Tilde{\bm X}_a^y})\frac{\mid\textup{MMD}(\bm X,\Tilde{\bm Y}) + (n_x + n_y)\mid}{n_x + n_y} + 2n_x\\
\le & \textup{Cond}(\bm{K}_{\Tilde{\bm X}_a^y,\Tilde{\bm X}_a^y})\left(\frac{\mid\textup{MMD}(\bm X,\Tilde{\bm Y})\mid}{n_x + n_y} + 1\right) + 2n_x\\
\end{align*}

Analogously, we derive an upper bound on the approximation error in terms of the Frobenius norm.

For $T_4$, we have
\begin{align*}
&\left\Vert\widehat{\bm{H}}_{\bm X, \bm{X}|\Tilde{\bm Y}} - \bm{K}_{\bm X, \bm X}\right\Vert_F^2 = \left\Vert\Tilde{\bm C}\Tilde{\bm W}_k^\dagger\Tilde{\bm C}^T - \bm{K}_{\Tilde{\bm X}_a^y, \Tilde{\bm X}_a^y}\right\Vert_F^2\\
\le & \left\Vert\bm{K}_{\Tilde{\bm X}_a^y, \Tilde{\bm X}_a^y} - \Bar{\bm{K}}_{\Tilde{\bm X}_a^y,\Tilde{\bm X}_a^y}\right\Vert_F^2 + 4\sqrt{k}\left\Vert \bm A\bm A^T \bm A\bm A^T - \bm C_{\bm A}\bm C_{\bm A}^T \bm C_{\bm A}\bm C_{\bm A}^T\right\Vert_F\\
= & \left\Vert\bm{K}_{\Tilde{\bm X}_a^y, \Tilde{\bm X}_a^y} - \Bar{\bm{K}}_{\Tilde{\bm X}_a^y,\Tilde{\bm X}_a^y}\right\Vert_F^2 + 4\sqrt{k}\left\Vert \bm A\bm A^T \bm A\bm A^T - \bm A\bm A^T \bm C_{\bm A}\bm C_{\bm A}^T + \bm A\bm A^T \bm C_{\bm A}\bm C_{\bm A}^T - \bm C_{\bm A}\bm C_{\bm A}^T \bm C_{\bm A}\bm C_{\bm A}^T\right\Vert_F\\
\le & \left\Vert\bm{K}_{\Tilde{\bm X}_a^y, \Tilde{\bm X}_a^y} - \Bar{\bm{K}}_{\Tilde{\bm X}_a^y,\Tilde{\bm X}_a^y}\right\Vert_F^2 + 4\sqrt{k}\left\Vert \bm A\bm A^T\left[\bm A\bm A^T - \bm C_{\bm A}\bm C_{\bm A}^T\right] + \left[\bm A\bm A^T - \bm C_{\bm A}\bm C_{\bm A}^T\right] \bm C_{\bm A}\bm C_{\bm A}^T\right\Vert_F\\
\le & \left\Vert\bm{K}_{\Tilde{\bm X}_a^y, \Tilde{\bm X}_a^y} - \Bar{\bm{K}}_{\Tilde{\bm X}_a^y,\Tilde{\bm X}_a^y}\right\Vert_F^2 + 4\sqrt{k}\left\{\left\Vert \bm A\bm A^T\right\Vert_F\left\Vert \bm A\bm A^T - \bm C_{\bm A}\bm C_{\bm A}^T\right\Vert_2 + \left\Vert \bm A\bm A^T - \bm C_{\bm A}\bm C_{\bm A}^T\right\Vert_2\left\Vert \bm C_{\bm A}\bm C_{\bm A}^T\right\Vert_F\right\}\\
= & \left\Vert\bm{K}_{\Tilde{\bm X}_a^y, \Tilde{\bm X}_a^y} - \Bar{\bm{K}}_{\Tilde{\bm X}_a^y,\Tilde{\bm X}_a^y}\right\Vert_F^2 + 4\sqrt{k}\left(\left\Vert \bm A\bm A^T\right\Vert_F + \left\Vert \bm C_{\bm A}\bm C_{\bm A}^T\right\Vert_F\right)\left\Vert \bm A\bm A^T - \bm C_{\bm A}\bm C_{\bm A}^T\right\Vert_2\\
= & \left\Vert\bm{K}_{\Tilde{\bm X}_a^y, \Tilde{\bm X}_a^y} - \Bar{\bm{K}}_{\Tilde{\bm X}_a^y,\Tilde{\bm X}_a^y}\right\Vert_F^2 + 4\sqrt{k}\left(\left\Vert \bm K_{\Tilde{\bm X}_a^y,\Tilde{\bm X}_a^y}\right\Vert_F + \left\Vert \bm K_{\Tilde{\bm Y},\Tilde{\bm Y}}\right\Vert_F\right)\left\Vert \bm A\bm A^T - \bm C_{\bm A}\bm C_{\bm A}^T\right\Vert_2\\
= & \sum_{i > k}\sigma_i^2\left(\bm{K}_{\Tilde{\bm X}_a^y, \Tilde{\bm X}_a^y}\right) + 4\sqrt{k}(n_x + 2n_y)n_x\\
\le & \sqrt{\sum_{i > k}\sigma_i^2\left(\bm{K}_{\Tilde{\bm X}_a^y, \Tilde{\bm X}_a^y}\right)} + 2k^{1/4}n_x\sqrt{1 + \frac{n_y}{n_x}}\\
\le & \sqrt{(n_x + n_y - k)\sigma_1^2\left(\bm{K}_{\Tilde{\bm X}_a^y, \Tilde{\bm X}_a^y}\right)} + 2k^{1/4}n_x\sqrt{1 + \frac{n_y}{n_x}}\\
= & \sqrt{n_x + n_y - k}\frac{\sigma_1\left(\bm{K}_{\Tilde{\bm X}_a^y, \Tilde{\bm X}_a^y}\right)}{\sigma_{n_x + n_y}\left(\bm{K}_{\Tilde{\bm X}_a^y, \Tilde{\bm X}_a^y}\right)}\sigma_{n_x + n_y}\left(\bm{K}_{\Tilde{\bm X}_a^y, \Tilde{\bm X}_a^y}\right) + 2k^{1/4}n_x\sqrt{1 + \frac{n_y}{n_x}}\\
\le & \sqrt{n_x + n_y - k}\textup{Cond}\left(\bm{K}_{\Tilde{\bm X}_a^y,\Tilde{\bm X}_a^y}\right)\frac{\mid\bm I_*^T\bm{K}_{\Tilde{\bm X}_a^y,\Tilde{\bm X}_a^y}\bm I_*\mid}{\bm I_*^T\bm I_*} + 2k^{1/4}n_x\sqrt{1 + \frac{n_y}{n_x}}\\
\le & \sqrt{n_x + n_y - k}\textup{Cond}\left(\bm{K}_{\Tilde{\bm X}_a^y,\Tilde{\bm X}_a^y}\right)\frac{\mid\bm I_*^T\bm{K}_{\Tilde{\bm X}_a^y,\Tilde{\bm X}_a^y}\bm I_*\mid}{\bm I_*^T\bm I_*} + 2k^{1/4}n_x\sqrt{1 + \frac{n_y}{n_x}}\\
\le & \sqrt{n_x + n_y - k}\textup{Cond}\left(\bm{K}_{\Tilde{\bm X}_a^y,\Tilde{\bm X}_a^y}\right)\left(\frac{\mid\textup{MMD}(\bm X,\Tilde{\bm Y})\mid}{n_x + n_y} + 1\right) + 2k^{1/4}n_x\sqrt{1 + \frac{n_y}{n_x}}
\end{align*}

\end{proof}

\section{Proof of Theorem \ref{thm:dp2}}
\begin{proof}

In our FedDL, consider a one-step gradient descent at client $p$
\begin{align*}
\bm Y_p^{s + 1} \leftarrow \bm Y_p^s - \eta_s\nabla f_p\left(\bm Y_p^s\right)
\end{align*}
where the derivative is given by
\begin{align*}
&\nabla f_p(\bm Y_p^s) = \frac{-4\gamma}{n_p n_y}\left[\bm X_p\bm{K}_{\bm X_p,\bm Y_p^s} - \bm Y_p^s\textup{Diag}(\bm 1_{n_p}^T\bm{K}_{\bm X_p,\bm Y_p^s})\right] + \frac{4\gamma}{n_y(n_y-1)}\left[\bm Y_p^s\bm{K}_{\bm Y_p^s,\bm Y_p^s} - \bm Y_p^s\textup{Diag}(\bm 1_{n_y}^T\bm{K}_{\bm Y_p^s,\bm Y_p^s})\right]
\end{align*}
with $\bm X_p = [(\bm X_p)_{:,1},\cdots,(\bm X_p)_{:,j-1},(\bm X_p)_{:,j},(\bm X_p)_{:,j+1},\cdots,(\bm X_p)_{:,n_p}]$.

In order to figure out the sensitivity of $g_{\bm Y_p}(\bm X_p) = \nabla f_p(\bm Y_p)$, we present the counterpart of the above expression with the neighboring data $\bm X_p'$ which differs only in one column from $\bm X_p$
\begin{align*}
& g_{\bm Y_p^s}(\bm X_p') = \frac{-4\gamma}{n_p n_y}\left[\bm X_p'\bm{K}_{\bm X_p',\bm Y_p^s} - \bm Y_p^s\textup{Diag}(\bm 1_{n_p}^T\bm{K}_{\bm X_p',\bm Y_p^s})\right] + \frac{4\gamma}{n_y(n_y-1)}\left[\bm Y_p^s\bm{K}_{\bm Y_p^s,\bm Y_p^s} - \bm Y_p^s\textup{Diag}(\bm 1_{n_y}^T\bm{K}_{\bm Y_p^s,\bm Y_p^s})\right]
\end{align*}
with $\bm X_p' = [(\bm X_p)_{:,1},\cdots,(\bm X_p)_{:,j-1},(\bm X_p')_{:,j},(\bm X_p)_{:,j+1},\cdots,(\bm X_p)_{:,n_p}]$.

Then, we derive an upper bound on the sensitivity of $g_{\bm Y_p}(\bm X_p) = \nabla f_p(\bm Y_p)$.
\begin{align*}
& \left\Vert g_{\bm Y_p^s}(\bm X_p) - g_{\bm Y_p^s}(\bm X_p')\right\Vert_F\\
= & \left\Vert\frac{-4\gamma}{n_pn_y}\left\{\left[\bm X_p\bm{K}_{\bm X_p,\bm Y_p^s} - \bm Y_p^s\textup{Diag}(\bm 1_{n_p}^T\bm{K}_{\bm X_p,\bm Y_p^s})\right] - \left[\bm X_p'\bm{K}_{\bm X_p',\bm Y_p^s} - \bm Y_p^s\textup{Diag}(\bm 1_{n_p}^T\bm{K}_{\bm X_p',\bm Y_p^s})\right]\right\}\right\Vert_F\\
= & \frac{4\gamma}{n_pn_y}\left\Vert\left[\bm X_p\bm{K}_{\bm X_p,\bm Y_p^s} - \bm X_p'\bm{K}_{\bm X_p',\bm Y_p^s}\right] - \left[\bm Y_p^s\textup{Diag}(\bm 1_{n_p}^T\bm{K}_{\bm X_p,\bm Y_p^s}) - \bm Y_p^s\textup{Diag}(\bm 1_{n_p}^T\bm{K}_{\bm X_p',\bm Y_p^s})\right]\right\Vert_F\\
\le & \frac{4\gamma}{n_pn_y}\left\{\underbrace{\left\Vert\bm X_p\bm{K}_{\bm X_p,\bm Y_p^s} - \bm X_p'\bm{K}_{\bm X_p',\bm Y_p^s}\right\Vert_F}_{T_1} + \underbrace{\left\Vert\bm Y_p^s\textup{Diag}(\bm 1_{n_p}^T(\bm{K}_{\bm X_p,\bm Y_p^s} - \bm{K}_{\bm X_p',\bm Y_p^s}))\right\Vert_F}_{T_2}\right\}
\end{align*}

We decompose $T_1$ into two parts by adding an intermediate term $\bm X_p'\bm{K}_{\bm X_p,\bm Y_p^s}$ and apply the triangle inequality to it.
\begin{align*}
& \left\Vert\bm X_p\bm{K}_{\bm X_p,\bm Y_p^s} - \bm X_p'\bm{K}_{\bm X_p',\bm Y_p^s}\right\Vert_F\\
\le & \underbrace{\left\Vert\bm X_p\bm{K}_{\bm X_p,\bm Y_p^s} - \bm X_p'\bm{K}_{\bm X_p,\bm Y_p^s}\right\Vert_F}_{T_3} + \underbrace{\left\Vert\bm X_p'\bm{K}_{\bm X_p,\bm Y_p^s} - \bm X_p'\bm{K}_{\bm X_p',\bm Y_p^s}\right\Vert_F}_{T_4}
\end{align*}

For $T_3$, we have
\begin{align*}
& \left\Vert\bm X_p\bm{K}_{\bm X_p,\bm Y_p^s} - \bm X_p'\bm{K}_{\bm X_p,\bm Y_p^s}\right\Vert_F\\
= & \left\Vert(\bm X_p - \bm X_p')\bm{K}_{\bm X_p,\bm Y_p^s}\right\Vert_F\\
= & \left\Vert((\bm X_p)_{:,j} - (\bm X_p')_{:,j})\bm{K}_{(\bm X_p)_{:,j},\bm Y_p^s}\right\Vert_2\\
\le & \left\Vert((\bm X_p)_{:,j} - (\bm X_p')_{:,j})\right\Vert_2\left\Vert\bm{K}_{(\bm X_p)_{:,j},\bm Y_p^s}\right\Vert_2\\
\le & \sqrt{n_y}\left\Vert(\bm X_p)_{:,j} - (\bm X_p')_{:,j}\right\Vert_2
\end{align*}
where we used the fact that $\left\Vert\bm{K}_{(\bm X_p)_{:,j},\bm Y_p^s}\right\Vert_2 \le \sqrt{n_y}$.

For $T_4$, we have
\begin{align*}
& \left\Vert\bm X_p'\bm{K}_{\bm X_p,\bm Y_p^s} - \bm X_p'\bm{K}_{\bm X_p',\bm Y_p^s}\right\Vert_F\\
= & \left\Vert\bm X_p'(\bm{K}_{\bm X_p,\bm Y_p^s} - \bm{K}_{\bm X_p',\bm Y_p^s})\right\Vert_F\\
= & \left\Vert(\bm X_p')_{:,j}(\bm{K}_{(\bm X_p)_{:,j},\bm Y_p^s} - \bm{K}_{(\bm X_p')_{:,j},\bm Y_p^s})\right\Vert_2\\
\le & \left\Vert(\bm X_p')_{:,j}\right\Vert_2\underbrace{\left\Vert\bm{K}_{(\bm X_p)_{:,j},\bm Y_p^s} - \bm{K}_{(\bm X_p')_{:,j},\bm Y_p^s}\right\Vert_2}_{T_5}
\end{align*}

Since $f(x) = \exp(x)$ is locally Lipschitz continuous when $x<0$, we have for $T_5$,
\begin{align*}
& \left\Vert\bm{K}_{(\bm X_p)_{:,j},\bm Y_p^s} - \bm{K}_{(\bm X_p')_{:,j},\bm Y_p^s}\right\Vert_2\\
= & \sqrt{\sum_{i = 1}^{n_y}\left(\bm{K}_{(\bm X_p)_{:,j},(\bm Y_p^s)_{:,i}} - \bm{K}_{(\bm X_p')_{:,j},(\bm Y_p^s)_{:,i}}\right)^2}\\
\le & \sqrt{\sum_{i = 1}^{n_y}\gamma^2\left(\left\Vert(\bm X_p)_{:,j} - (\bm Y_p^s)_{:,i}\right\Vert_2^2 - \left\Vert(\bm X_p')_{:,j} - (\bm Y_p^s)_{:,i}\right\Vert_2^2\right)^2}\\
= & \left\{\sum_{i = 1}^{n_y}\gamma^2\left(\left\Vert(\bm X_p)_{:,j} - (\bm X_p')_{:,j}\right\Vert_2^2 + 2\left\langle(\bm Y_p^{k - 1})_{:,i} - (\bm X_p')_{:,j},(\bm X_p')_{:,j} - (\bm X_p)_{:,j}\right\rangle\right)^2\right\}^{1/2}\\
\le & \left\{\sum_{i = 1}^{n_y}\gamma^2\left(\left\Vert(\bm X_p)_{:,j} - (\bm X_p')_{:,j}\right\Vert_2^2 + 2\left\Vert(\bm Y_p^{k - 1})_{:,i} - (\bm X_p')_{:,j}\right\Vert_2\left\Vert(\bm X_p')_{:,j} - (\bm X_p)_{:,j}\right\Vert_2\right)^2\right\}^{1/2}\\
\end{align*}

Assume $\max_{p,j}\Vert(\bm X_p)_{:,j}\Vert_2 = \tau_X$, $\max_{p,i,j}\Vert(\bm Y_{p})_{:,i} - (\bm X_p)_{:,j}\Vert = \Upsilon$, we thus get an upper bound on $T_1$
\begin{align*}
& \left\Vert\bm X_p\bm{K}_{\bm X_p,\bm Y_p^s} - \bm X_p'\bm{K}_{\bm X_p',\bm Y_p^s}\right\Vert_F\\
\le & 2\sqrt{n_y}\tau_X + \tau_X\sqrt{\sum_{i = 1}^{n_y}\gamma^2\left(4\tau_X^2 + 4\Upsilon\tau_X\right)^2}\\
\le & 2\sqrt{n_y}\tau_X + 4\sqrt{n_y}\gamma\tau_X^2\left(\tau_X + \Upsilon\right)\\
= & 2\sqrt{n_y}\tau_X\left(1 + 2\gamma\tau_X\left(\tau_X + \Upsilon\right)\right)
\end{align*}

Suppose $\Vert\bm Y_p^s\Vert_{sp} \le \tau_Y$, we have for $T_2$,
\begin{align*}
& \left\Vert\bm Y_p^s\textup{Diag}(\bm 1_{n_p}^T(\bm{K}_{\bm X_p,\bm Y_p^s} - \bm{K}_{\bm X_p',\bm Y_p^s}))\right\Vert_F\\
\le & \left\Vert\bm Y_p^s\right\Vert_{sp}\left\Vert\textup{Diag}(\bm 1_{n_p}^T(\bm{K}_{\bm X_p,\bm Y_p^s} - \bm{K}_{\bm X_p',\bm Y_p^s}))\right\Vert_F\\
= & \left\Vert\bm Y_p^s\right\Vert_{sp}\left\Vert\bm{K}_{(\bm X_p)_{:,j},\bm Y_p^s} - \bm{K}_{(\bm X_p')_{:,j},\bm Y_p^s}\right\Vert_2\\
\le & 4\sqrt{n_y}\gamma\tau_X\tau_Y\left(\tau_X + \Upsilon\right)\\
\end{align*}

Combined with the above conclusions, we give an upper bound on the $\ell_2$-sensitivity of $g_{\bm Y_p}(\bm X_p) = \nabla f_p(\bm Y_p)$.
\begin{align*}
& \triangle_2(g_{\bm Y_p}) = \sup_{\bm X_p\sim\bm X_p'}\left\Vert g_{\bm Y_p^s}(\bm X_p) - g_{\bm Y_p^s}(\bm X_p')\right\Vert_F\\
\le & \frac{4\gamma}{n_pn_y}\left\{2\sqrt{n_y}\tau_X\left(1 + 2\gamma\tau_X\left(\tau_X + \Upsilon\right)\right) + 4\sqrt{n_y}\gamma\tau_X\tau_Y\left(\tau_X + \Upsilon\right)\right\}\\
\le & \frac{4\gamma}{n_pn_y}\cdot 2\sqrt{n_y}\tau_X\left\{1 + 2\gamma(\tau_X + \tau_Y)\left(\tau_X + \Upsilon\right)\right\}\\
= & \frac{8\sqrt{n_y}\gamma\tau_X}{n_pn_y}\left\{1 + 2\gamma(\tau_X + \tau_Y)\left(\tau_X + \Upsilon\right)\right\}
\end{align*}

Assume $\Delta = \frac{8\sqrt{n_y}\gamma\tau_X}{n_pn_y}\left\{1 + 2\gamma(\tau_X + \tau_Y)\left(\tau_X + \Upsilon\right)\right\}$, the mechanism that injects Gaussian noise to $\nabla f_p(\bm Y_p^s)$ for $s\in[S]$ with variance $8S\Delta^2\log(e + (\varepsilon/\delta))/\varepsilon^2$ satisfies $(\varepsilon,\delta)-$differential privacy under $S-$fold adaptive composition for any $\varepsilon > 0$ and $\delta\in(0,1]$ by Theorem 4.3 of \cite{kairouz2015composition}.

\end{proof}

\section{Proof of Convergence of Algorithm \ref{alg:feddl}}
\label{prf: convergence of feddl}
\begin{proof}
Assume the gradient of all local objective functions $f_p(\cdot)$ for $p=1,\dots,P$ are $L_p$-Lipschitz continuous, that is, for all $\bm Y',\bm Y$ 
\begin{align*}
\left\Vert\nabla f_p(\bm Y') - \nabla f_p(\bm Y)\right\Vert_F \le L_p\left\Vert\bm Y' - \bm Y\right\Vert_F
\end{align*}
Then, we have the descent formula for client $p$
\begin{align*}
f_p(\bm Y') - f_p(\bm Y) \le & \langle\nabla f_p(\bm Y), \bm Y' - \bm Y\rangle + \frac{L_p}{2}\Vert\bm Y' - \bm Y\Vert_F^2
\end{align*}
With $\sum_{p=1}^P\omega_p=1$ and $L=\sum_{p=1}^P \omega_p L_p$, we have
\begin{align*}
& F(\bm Y^{s,t}) - F(\bm Y^{s,t-1}) = \sum_{p=1}^P\omega_p f_p(\bm Y^{s,t}) - \sum_{p=1}^P\omega_p f_p(\bm Y^{s,t-1})
= \sum_{p=1}^P\omega_p\left[f_p(\bm Y^{s,t}) - f_p(\bm Y^{s,t-1})\right]\\
\le & \sum_{p=1}^P\omega_p\left[\langle\nabla f_p(\bm Y^{s,t-1}), \bm Y^{s,t} - \bm Y^{s,t-1}\rangle + \frac{L_p}{2}\Vert\bm Y^{s,t} - \bm Y^{s,t-1}\Vert_F^2\right]\\
= & \langle\sum_{p=1}^P\omega_p\nabla f_p(\bm Y^{s,t-1}), \bm Y^{s,t} - \bm Y^{s,t-1}\rangle + \frac{\sum_{p=1}^P\omega_p L_p}{2}\Vert\bm Y^{s,t} - \bm Y^{s,t-1}\Vert_F^2\\
= & \langle\nabla F(\bm Y^{s,t-1}), \bm Y^{s,t} - \bm Y^{s,t-1}\rangle + \frac{L}{2}\Vert\bm Y^{s,t} - \bm Y^{s,t-1}\Vert_F^2\\
\end{align*}

When updating the local $\bm Y_p$ by gradient descent with the step size $1/L$
\begin{align*}
& \bm Y_p^{s,t} = \bm Y_p^{s,t-1} - \frac{1}{L}\nabla f_p(\bm Y_p^{s,t-1})\\
\iff & \bm Y^{s,t} = \bm Y^{s,t-1} - \frac{1}{L}\sum_{p=1}^P\omega_p\nabla f_p(\bm Y_p^{s,t-1})\\
\iff & \sum_{p=1}^P\omega_p\nabla f_p(\bm Y_p^{s,t-1}) + L(\bm Y^{s,t} - \bm Y^{s,t-1}) = 0
\end{align*}

Thus, 
\begin{align*}
& F(\bm Y^{s,t}) - F(\bm Y^{s,t-1})\\
\le & \langle\nabla F(\bm Y^{s,t-1}), \bm Y^{s,t} - \bm Y^{s,t-1}\rangle + \frac{L}{2}\Vert\bm Y^{s,t} - \bm Y^{s,t-1}\Vert_F^2\\
= & \langle\nabla F(\bm Y^{s,t-1}) - \sum_{p=1}^P\omega_p\nabla f_p(\bm Y_p^{s,t-1}) - L(\bm Y^{s,t} - \bm Y^{s,t-1}), \bm Y^{s,t} - \bm Y^{s,t-1}\rangle + \frac{L}{2}\Vert\bm Y^{s,t} - \bm Y^{s,t-1}\Vert_F^2\\
= & -\frac{L}{2}\Vert\bm Y^{s,t} - \bm Y^{s,t-1}\Vert_F^2 + \langle\nabla F(\bm Y^{s,t-1}) - \sum_{p=1}^P\omega_p\nabla f_p(\bm Y_p^{s,t-1}), \bm Y^{s,t} - \bm Y^{s,t-1}\rangle\\
= & -\frac{L}{4}\Vert\bm Y^{s,t} - \bm Y^{s,t-1}\Vert_F^2 + \frac{1}{L}\underbrace{\Vert\nabla F(\bm Y^{s,t-1}) - \sum_{p=1}^P\omega_p\nabla f_p(\bm Y_p^{s,t-1})\Vert_F^2}_{T_1}
\end{align*}

For $T_1$,
\begin{align*}
& \Vert\nabla F(\bm Y^{s,t-1}) - \sum_{p=1}^P\omega_p\nabla f_p(\bm Y_p^{s,t-1})\Vert_F^2\\
= & \Vert\sum_{p=1}^P\omega_p\nabla f_p(\bm Y^{s,t-1}) - \sum_{p=1}^P\omega_p\nabla f_p(\bm Y_p^{s,t-1})\Vert_F^2\\
= & \Vert\sum_{p=1}^P\omega_p\left[\nabla f_p(\bm Y^{s,t-1}) - \nabla f_p(\bm Y_p^{s,t-1})\right]\Vert_F^2\\
\le & \sum_{p=1}^P\omega_p\Vert\nabla f_p(\bm Y^{s,t-1}) - \nabla f_p(\bm Y_p^{s,t-1})\Vert_F^2\\
\le & \sum_{p=1}^P\omega_p L_p^2\underbrace{\Vert\bm Y^{s,t-1} - \bm Y_p^{s,t-1}\Vert_F^2}_{T_2}
\end{align*}
where we used the assumption for the last inequality that $\nabla f_p(\cdot)$ is $L$-Lipschitz continuous.

By Observing that
\begin{align*}
\begin{cases}
\bm Y_p^{s,t-1} = \bm Y_p^{s,0} - \frac{1}{L}\sum_{j=1}^{t-1}\nabla f_p(\bm Y_p^{s,j-1})\\
\bm Y^{s,t-1} = \bm Y^{s,0} - \frac{1}{L}\sum_{j=1}^{t-1}\sum_{{p'}=1}^P\omega_{p'}\nabla f_{p'}(\bm Y_{p'}^{s,j-1})\\
\bm Y_p^{s,0} = \bm Y^{s,0}
\end{cases},
\end{align*}
we have for $T_2$
\begin{align*}
& \Vert\bm Y^{s,t-1} - \bm Y_p^{s,t-1}\Vert_F^2\\
= & \Vert\frac{1}{L}\sum_{j=1}^{t-1}\nabla f_p(\bm Y_p^{s,j-1}) - \frac{1}{L}\sum_{j=1}^{t-1}\sum_{{p'}=1}^P\omega_{p'}\nabla f_{p'}(\bm Y_{p'}^{s,j-1})\Vert_F^2\\
\le & \frac{t-1}{L^2}\sum_{j=1}^{t-1}\Vert\nabla f_p(\bm Y_p^{s,j-1}) - \sum_{{p'}=1}^P\omega_{p'}\nabla f_{p'}(\bm Y_{p'}^{s,j-1})\Vert_F^2\\
= & \frac{t-1}{L^2}\sum_{j=1}^{t-1}\Vert\sum_{{p'}=1}^P\omega_{p'}[\nabla f_p(\bm Y_p^{s,j-1}) - \nabla f_{p'}(\bm Y_{p'}^{s,j-1})]\Vert_F^2\\
\le & \frac{t-1}{L^2}\sum_{j=1}^{t-1}\sum_{{p'}=1}^P\omega_{p'}\underbrace{\Vert\nabla f_p(\bm Y_p^{s,j-1}) - \nabla f_{p'}(\bm Y_{p'}^{s,j-1})\Vert_F^2}_{T_3}
\end{align*}

For $T_3$,
\begin{align*}
& \Vert\nabla f_p(\bm Y_p^{s,j-1}) - \nabla f_{p'}(\bm Y_{p'}^{s,j-1})\Vert_F^2\\
= & \Vert\nabla f_p(\bm Y_p^{s,j-1}) - \nabla f_p(\bm Y^{s,j-1}) + \nabla f_p(\bm Y^{s,j-1}) - \nabla f_{p'}(\bm Y_{p'}^{s,j-1}) + \nabla f_{p'}(\bm Y^{s,j-1}) - \nabla f_{p'}(\bm Y^{s,j-1})\Vert_F^2\\
= & \Vert[\nabla f_p(\bm Y_p^{s,j-1}) - \nabla f_p(\bm Y^{s,j-1})] + [\nabla f_{p'}(\bm Y^{s,j-1}) - \nabla f_{p'}(\bm Y_{p'}^{s,j-1})] + [\nabla f_p(\bm Y^{s,j-1}) - \nabla f_{p'}(\bm Y^{s,j-1})]\Vert_F^2\\
\le & 3\Vert\nabla f_p(\bm Y_p^{s,j-1}) - \nabla f_p(\bm Y^{s,j-1})\Vert_F^2 + 3\Vert\nabla f_{p'}(\bm Y^{s,j-1}) - \nabla f_{p'}(\bm Y_{p'}^{s,j-1})\Vert_F^2 + 3\Vert\nabla f_p(\bm Y^{s,j-1}) - \nabla f_{p'}(\bm Y^{s,j-1})\Vert_F^2\\
\le & 3L_p^2\Vert\bm Y_p^{s,j-1} - \bm Y^{s,j-1}\Vert_F^2 + 3L_{p'}^2\Vert\bm Y_{p'}^{s,j-1} - \bm Y^{s,j-1}\Vert_F^2 + 3\zeta^2
\end{align*}

Thus, we have for $T_2$,
\begin{align*}
& \Vert\bm Y^{s,t-1} - \bm Y_p^{s,t-1}\Vert_F^2\\
\le & \frac{t-1}{L^2}\sum_{j=1}^{t-1}\sum_{{p'}=1}^P\omega_{p'}\underbrace{\Vert\nabla f_p(\bm Y_p^{s,j-1}) - \nabla f_{p'}(\bm Y_{p'}^{s,j-1})\Vert_F^2}_{T_3}\\
\le & \frac{t-1}{L^2}\sum_{j=1}^{t-1}\sum_{{p'}=1}^P\omega_{p'}\left[3L_p^2\Vert\bm Y_p^{s,j-1} - \bm Y^{s,j-1}\Vert_F^2 + 3L_{p'}^2\Vert\bm Y_{p'}^{s,j-1} - \bm Y^{s,j-1}\Vert_F^2 + 3\zeta^2\right]\\
= & \frac{3t^2\zeta^2}{L^2} + \frac{3(t-1)}{L^2}\sum_{j=1}^{t-1}L_p^2\Vert\bm Y_p^{s,j-1} - \bm Y^{s,j-1}\Vert_F^2 + \frac{3(t-1)}{L^2}\sum_{j=1}^{t-1}\sum_{{p'}=1}^P\omega_{p'}L_{p'}^2\Vert\bm Y_{p'}^{s,j-1} - \bm Y^{s,j-1}\Vert_F^2
\end{align*}

With $\frac{\sum_{p=1}^P\omega_p L_p^2}{L^2} = \frac{\sum_{p=1}^P\omega_p L_p^2}{(\sum_{p=1}^P\omega_p L_p)^2} = \rho_L$, we further have,
\begin{align*}
& \sum_{t=1}^Q\sum_{p=1}^P\omega_p L_p^2\underbrace{\Vert\bm Y^{s,t-1} - \bm Y_p^{s,t-1}\Vert_F^2}_{T_2}\\
\le & \sum_{t=1}^Q\sum_{p=1}^P\omega_p L_p^2\left[\frac{3t^2\zeta^2}{L^2} + \frac{3(t-1)}{L^2}\sum_{j=1}^{t-1}L_p^2\Vert\bm Y_p^{s,j-1} - \bm Y^{s,j-1}\Vert_F^2 + \frac{3(t-1)}{L^2}\sum_{j=1}^{t-1}\sum_{{p'}=1}^P\omega_{p'}L_{p'}^2\Vert\bm Y_{p'}^{s,j-1} - \bm Y^{s,j-1}\Vert_F^2\right]\\
= & 3\rho_L\zeta^2\sum_{t=1}^Q t^2 + \sum_{t=1}^Q\sum_{p=1}^P\omega_p L_p^2\frac{3(t-1)}{L^2}\sum_{j=1}^{t-1}L_p^2\Vert\bm Y_p^{s,j-1} - \bm Y^{s,j-1}\Vert_F^2 + \sum_{t=1}^Q\sum_{p=1}^P\omega_p L_p^2\frac{3(t-1)}{L^2}\sum_{j=1}^{t-1}\sum_{{p'}=1}^P\omega_{p'}L_{p'}^2\Vert\bm Y_{p'}^{s,j-1} - \bm Y^{s,j-1}\Vert_F^2\\
= & 3\rho_L\zeta^2\sum_{t=1}^Q t^2 + \sum_{t=1}^Q 3(t-1)\sum_{j=1}^{t-1}\sum_{p=1}^P\omega_p \frac{L_p^2}{L^2} L_p^2\Vert\bm Y_p^{s,j-1} - \bm Y^{s,j-1}\Vert_F^2 + \sum_{t=1}^Q 3(t-1)\sum_{j=1}^{t-1}\sum_{{p}=1}^P\omega_{p}\rho_L L_{p}^2\Vert\bm Y_{p}^{s,j-1} - \bm Y^{s,j-1}\Vert_F^2\\
= & 3\rho_L\zeta^2\sum_{t=1}^Q t^2 + \sum_{t=1}^Q 3(t-1)\sum_{j=1}^{t-1}\sum_{p=1}^P\omega_p (\rho_L + \frac{L_p^2}{L^2}) L_p^2\Vert\bm Y_p^{s,j-1} - \bm Y^{s,j-1}\Vert_F^2\\
\le & 3\rho_L\zeta^2 Q(Q+1)(2Q+1) + 3(Q-1)^2(\rho_L + \frac{\max_p L_p^2}{L^2})\sum_{t=1}^Q\sum_{p=1}^P\omega_p L_p^2\Vert\bm Y_p^{s,t-1} - \bm Y^{s,t-1}\Vert_F^2
\end{align*}

Thus, we can get
\begin{align*}
& \sum_{t=1}^Q\sum_{p=1}^P\omega_p L_p^2\underbrace{\Vert\bm Y^{s,t-1} - \bm Y_p^{s,t-1}\Vert_F^2}_{T_2} \le \frac{3\rho_L\zeta^2 Q(Q+1)(2Q+1)}{1 - 3(Q-1)^2(\rho_L + \frac{\max_p L_p^2}{L^2})}
\end{align*}

Consequently, we have
\begin{align*}
& F(\bm Y^s) - F(\bm Y^{s-1}) = F(\bm Y^{s,Q}) - F(\bm Y^{s,0}) = \sum_{t=1}^Q F(\bm Y^{s,t}) - F(\bm Y^{s,t-1})\\
= & -\frac{L}{4}\Vert\bm Y^{s,t} - \bm Y^{s,t-1}\Vert_F^2 + \frac{1}{L}\underbrace{\Vert\nabla F(\bm Y^{s,t-1}) - \sum_{p=1}^P\omega_p\nabla f_p(\bm Y_p^{s,t-1})\Vert_F^2}_{T_1}\\
= & -\frac{L}{4}\sum_{t=1}^Q\Vert\bm Y^{s,t} - \bm Y^{s,t-1}\Vert_F^2 + \frac{1}{L}\sum_{t=1}^Q\sum_{p=1}^P\omega_p L_p^2\underbrace{\Vert\bm Y^{s,t-1} - \bm Y_p^{s,t-1}\Vert_F^2}_{T_2}\\
\le & -\frac{L}{4}\sum_{t=1}^Q\Vert\bm Y^{s,t} - \bm Y^{s,t-1}\Vert_F^2 + \frac{3\rho_L\zeta^2 Q(Q+1)(2Q+1)}{L[1 - 3(Q-1)^2(\rho_L + \frac{\max_p L_p^2}{L^2})]}\\
\end{align*}

Thus,
\begin{align*}
& \sum_{t=1}^Q\Vert\bm Y^{s,t} - \bm Y^{s,t-1}\Vert_F^2 \le \frac{4}{L}[F(\bm Y^{s-1}) - F(\bm Y^{s})] + \frac{12\rho_L\zeta^2 Q(Q+1)(2Q+1)}{L^2[1 - 3(Q-1)^2(\rho_L + \frac{\max_p L_p^2}{L^2})]}\\
\end{align*}

Summing it from $s=1$ to $S$ followed by the average, we immediately get
\begin{align*}
& \frac{1}{SQ}\sum_{s=1}^S\sum_{t=1}^Q\Vert\bm Y^{s,t} - \bm Y^{s,t-1}\Vert_F^2 \le \frac{4}{SQL}[F(\bm Y^0) - F(\bm Y^S)] + \frac{12\rho_L\zeta^2 (Q+1)(2Q+1)}{L^2[1 - 3(Q-1)^2(\rho_L + \frac{\max_p L_p^2}{L^2})]}\\
\end{align*}
\end{proof}

\end{document}